\documentclass[lettersize,journal]{IEEEtran}
\usepackage{amssymb}
\usepackage{pifont}
\usepackage{amsmath,amsfonts}
\usepackage{algorithmic}
\usepackage{algorithm}
\usepackage{array}
\usepackage[caption=false,font=normalsize,labelfont=sf,textfont=sf]{subfig}
\usepackage{textcomp}
\usepackage{stfloats}
\usepackage{url}
\usepackage{verbatim}
\usepackage{graphicx}
\usepackage[table]{xcolor}
\usepackage{cite}
\usepackage{multirow}
\usepackage{multicol} % 导言区添加
\usepackage{adjustbox}
\usepackage{caption}
\usepackage{booktabs}
\usepackage{lmodern}
\usepackage{algorithm}
\usepackage{algorithmic}
\usepackage{subfig}
\usepackage{tabularx}
\usepackage{subcaption}
\usepackage{bm}
\usepackage{longtable}
\usepackage{array}
\usepackage{multirow}
\usepackage{bbding}
\usepackage{float}
\usepackage{multirow}
\usepackage{bbding}
\usepackage{booktabs}
\begin{document}
\title{Adaptive Disentangled Representation Learning for Incomplete Multi-View Multi-Label Classification}

\author{ Quanjiang~Li$^{\dag}$, Zhiming~Liu$^{\dag}$,   Tianxiang~Xu$^{\dag}$, Tingjin~Luo$^{*}$ and Chenping~Hou,~\IEEEmembership{Member,~IEEE}
        % <-this % stops a space
\thanks{ Quanjiang Li, Tingjin Luo and   Chenping Hou are with the
College of Science, National University of Defense Technology, Changsha
410073, China. E-mail: liquanjiang@nudt.edu.cn,
tingjinluo@hotmail.com, 
hcpnudt@hotmail.com.   }% <-this % 
\thanks{ Zhiming Liu is with College of Artificial Intelligence, Harbin Institute of Technology, Shenzhen, 518000, China. E-mail: mikuzliu@gmail.com.}
\thanks{ Tianxiang Xu is with School of Software and Microelectronics, Peking University, Beijing, 102600, China. E-mail: xtx\_pku@stu.pku.edu.cn.}

\thanks{$^{\dag}$These authors contributed equally to this work.}
\thanks{$^{*}$Corresponding author.}
}



% \IEEEpubid{0000}

\maketitle
\begin{abstract}
Multi-view multi-label learning frequently suffers from simultaneous feature absence and incomplete annotations, due to challenges in data acquisition and cost-intensive supervision. To tackle the complex yet highly practical problem while overcoming the existing limitations of feature recovery, representation disentanglement, and  label semantics modeling, we propose an Adaptive Disentangled Representation Learning method (ADRL). ADRL achieves robust view completion by propagating feature-level affinity across modalities with neighborhood awareness, and reinforces reconstruction effectiveness by leveraging a stochastic masking strategy. Through disseminating category-level association across label distributions, ADRL refines  distribution parameters for capturing interdependent label prototypes. Besides,  we formulate a mutual-information–based objective to promote consistency among shared representations and suppress information overlap between view-specific representation and other modalities.  Theoretically, we derive the tractable bounds to train the dual-channel network. Moreover, ADRL performs prototype-specific feature selection by enabling independent interactions between label embeddings and view representations, accompanied by the generation of pseudo-labels for each category.  The structural characteristics of the pseudo-label space are then exploited to  guide a discriminative trade-off during view fusion.  Finally, extensive experiments on public datasets and real-world applications demonstrate the superior performance of ADRL.
\end{abstract}

\section{Introduction}
\IEEEPARstart{T}{he} value of integrating information from diverse sources lies in its capacity to offer deeper insights and a more comprehensive understanding of issues. By overcoming individual limitations and bridging data gaps through data combination \cite{chen2019learning,xu2022self}, multi-view learning has emerged as an exceptionally promising domain. With the exponential expansion of Multimodal data, it holds significant potential to elevate performance across a wide array of conventional tasks \cite{wen2022survey,fang2021animc,fang2022multi}. Meanwhile, multi-label classification, where each instance can be associated with multiple relevant labels, inherently correspond to the  accelerated growth of annotation quantity in the information age. 
 For example, in image recognition \cite{sun2024deep}, a single image may be labeled with terms such as ``cat'', ``grass'' and ``outdoor'',  while in text classification \cite{chang2020taming}, a document might simultaneously fall under categories like ``politics'',  ``economics'' and ``international relations''.  The  identification of multi-label frequently relies on the support of rich feature pool. Therefore, multi-view data is particularly well-suited for multi-label classification, as multi-view data inherently provides the necessary diversity and complementary perspectives and effectively fulfills the requirement of distinct feature semantics for predicting high-dimension label space \cite{hao2025uncertainty}.     Moreover, a synthesis of diverse perspectives and thorough labeling is fundamental to accurately describe the complex data of the real world. Overall, this work focuses on the multi-view
multi-label classification (MvMLC) task,  which demonstrates greater potential in practical applications. \cite{liu2025reliable,zhang2018latent}.

Existing MvMLC methods can perform the joint analysis of diverse features and the collaborative recognition of multiple objectives, such as lrMMC \cite{liu2015low}  based on low-rank matrix factorization and E$F^{2}$FS \cite{hao2025embedded} grounded in feature selection. However,  these methods hinge on the premise that complete information from all views and labels is accessible,  which is evidently difficult to satisfy in practice. Due to the challenges associated with feature acquisition and the inconvenience posed by data storage, multi-view data often suffers from incompleteness. For example, in remote sensing, a region might be covered by multispectral satellite imagery but miss hyperspectral or LiDAR data because of sensor limitations and high storage demands \cite{SEC-LSRM}. In addition, owing to the prohibitive cost of manual labeling and the intrinsic semantic ambiguity of certain categories, multi-label data frequently contain only partial label information.  In medical imaging, a chest X-ray may involve multiple conditions, yet only a limited set is annotated as  expert labeling is resource-intensive, sensitive data must be protected, and certain findings lack clear diagnostic boundaries \cite{sun2024fedmlp}. Therefore, the simultaneous absence of views and labels constitutes a realistic and prevalent issue that warrants further investigation. The complexity arising from the diversity of features and labels, coupled with the scarcity of informative content, makes tackling incomplete multi-view multi-label  classification (iMvMLC) tasks particularly challenging.  

Despite the emergence of many effective iMvMLC methods, they still remain inadequate in several aspects, including  feature reconstruction,  semantic clarity of extracted representations, and the modeling of label embeddings enriched with correlation information, as well as their coupling with feature representations. (i) The absence of views severely  impairs the learning of expressive representations and compromises the stability of downstream modules. Many approaches merely employ a missing indicator to mask out unavailable views. This tendency to sidestep the issue of view incompleteness, rather than  exploring robust reconstruction strategies, inevitably leads to performance bottlenecks.  AIMNet \cite{liu2024attention} explored missing imputation through an instance-level global attention propagation mechanism. Nonetheless, relying on weighted contributions from all available instances to recover missing entries proves unreliable, as it tends to exacerbate reconstruction noise due to the limited relevance of many samples. (ii) It is widely acknowledged that the essence of multi-view representation learning lies in capturing shared semantics across views while preserving view-specific discriminative features. Most researches emphasize learning consistent representations and neglect the preservation of view-specific information, which is unfavorable for multi-label learning, as certain labels depend strongly on distinctive cues from individual views. MTD \cite{liu2023masked} employed geometric distance constraints to disentangle shared and specific representations, yet its linear interaction scheme fell short in capturing the complex inter-view dependencies. 
Effectively modeling intrinsic interactions at the representation level and achieving feature separation remains a persistent challenge.
(iii)
Label correlation plays a pivotal role in multi-label learning and distinguishes it from the multi-class setting, which renders approaches based on independent binary classification \cite{xiao2024new} insufficient for predicting multiple labels. Rather than treating interdependent label semantics in isolation, they should interact seamlessly with the representation information of multi-view to enhance the identification of label-specific features.  Moreover, view fusion should account not only for the feature-level information but also for the structural characteristics embedded in the multi-label space.

To address these problems, we propose an  Adaptive Disentangled Representation Learning method  for iMvMLC named ADRL. The central motivation of ADRL is to reduce reconstruction discrepancy,  promote the extraction of  semantically discriminative representations and strengthen the alignment and interaction between the feature and label space. Firstly, we construct view-specific instance-level affinity graphs  by diffusing attention-induced sample similarity across views. Then, neighborhood-aware selection and relevance-weighted aggregation are applied for view completion. The entire process is data-driven and free from trainable parameters, thereby reducing the potential source of error. Besides, a stochastic fragment masking strategy is employed to filter out low-quality reconstructed features and enhance the expressive stability  of feature semantics. Next, we design a mutual information-based model to disentangle the representation information, which not only emphasizes semantic consistency between the learned representation and the original view, but also enforces interactions among shared representations, while suppressing redundancy among private features.  Moreover, grounded in information-theoretic principles, we derive the tractable mutual information bound that serves as the training objective. Furthermore, a graph attention network is employed to propagate inter-label dependencies and update label distribution parameters. Subsequently, sampling from the refined distribution yields correlated label embeddings. Each label prototype interacts with shared and private representations to generate label-specific features and corresponding pseudo-labels. Then, these pseudo-labels  are paired with view representations to compute a manifold consistency loss.  The inverse of this loss serves as a fusion weight, assigning greater importance to views that exhibit stronger structural alignment with the label space. Ultimately,  the shared representation is enhanced with view-specific information to produce a unified feature for final prediction. The main contributions are summarized as below:
\begin{itemize}
\item[$\bullet$]
We propose a general multi-view representation learning framework based on mutual information modeling, which adaptively mitigates information shift of the learned representations and promotes effective information disentanglement. Theoretically, we derive the reliable bounds
of mutual information suitable for optimization to guide training.

\item[$\bullet$] ADRL enhances the propagation of feature associations and label correlations
to efficiently recover missing views and strengthen the semantic learning of label embeddings, respectively.
Meanwhile, feature selection is applied to label prototypes, with label information further integrated into view fusion.

\item[$\bullet$] Extensive experiments on diverse public datasets, together with applications to real-world NBA data, demonstrate the effectiveness of ADRL.
\end{itemize}

\section{Related Works}
\subsection{MvMLC}
Although the integration of multi-view learning with multi-label classification inevitably introduces  algorithmic complexity, it enables an exhaustive characterization of object attributes. Then, the algorithm is supposed to be designed to handle both multi-view processing and multi-label recognition  simultaneously. For instance, a method called LSA-MML \cite{zhang2018latent} learned a shared representation via matrix factorization and aligned latent basis matrices of different views in the kernel space using the Hilbert-Schmidt Independence Criterion (HSIC) technique, with the objective of maximizing the inter-view dependence. Wu et al. introduced SIMM, a deep MvMLC network that focused on jointly learning a shared subspace and view-specific features \cite{wu2019multi}. A central mechanism in SIMM involved enforcing a high degree of distinction between the shared and specific representation, allowing the network to capture the unique contribution of each view. Unlike LSA-MML and SIMM, CDMM \cite{zhao2021consistency} prioritizes the exploration of label relevance. It aims to  pursue inter-view consistency and diversity through a simple yet effective approach, and models label correlations by constructing a label affinity matrix that integrates Jaccard similarity in the label space with instance-level proximity in the feature space, followed by label propagation to enhance label representation. LVSL \cite{zhao2022non} is a sophisticated method for non-aligned multi-view multi-label classification that jointly explored view-specific labels and low-rank label structures while preserving the geometric characteristics of the original data via Laplacian graph regularization.

\subsection{iMvMLC}
Incomplete multi-view multi-label learning has emerged as a prominent research topic and experienced rapid advancements. This challenging problem encompasses missing data patterns across the domains of feature information and supervision signals. Missing views are typically handled by either ignoring them with prior guidance or reconstructing them via data completion. For instance, in incomplete multi-view clustering, Liu et al. \cite{liu2022localized} introduced a missing indicator matrix that enabled the model to adaptively accommodate instances with absent views. For missing view imputation, Wen et al.~\cite{wen2019unified} proposed UEAF, which recovered missing views using a  error matrix informed by local structure and enhanced alignment via reverse graph learning. The issue of incomplete labels in multi-label learning has stimulated the emergence of numerous methods. DM2L~\cite{ma2021expand} inferred latent labels from partial annotations, integrated label correlation and low-rank constraints for robust feature learning, and refined pseudo-labels via self-supervision learning. Another representative single-view multi-label classification model GLOCAL \cite{zhu2017multi} constructed bidirectional graphs between labels and instances to capture global label semantics under missing annotations. It further incorporated local priors and structure-preserving constraints to mitigate label noise and enhance feature discriminability.

Traditional methods like iMVWL addressed the simultaneous missingness of both  views and labels through jointly learning a shared subspace, weak-label predictor, and low-rank label correlation matrix, with their predictive performance constrained by the limited ability to capture shallow representations \cite{tan2018incomplete}. Building on the remarkable capabilities of deep learning, an increasing number of advanced methods have been developed, which leads to substantial improvements in performance. The initial exploration DIMC \cite{wen2023deep} was built upon deep neural networks with an end-to-end architecture to extract high-level discriminative features, which  further enhanced representation robustness via decoder-based reconstruction.  Then, DICNet  \cite{liu2023dicnet} centered on instance-level contrastive learning that pulled together representations of the same sample across views while pushing apart those of different instances, thereby enhancing the expressive power of common information.   LMVCAT \cite{liu2023incomplete}  proposed a Transformer-based framework, where the  feature extraction module leveraged multi-head attention pattern to enable feature interactions and incorporated label information to mitigate  structural distortions in the feature space, and the category-aware module captured the label  semantic information by facilitating the interplay of subcategory embeddings. Moreover, several efforts have been made to investigate alternative effective mechanisms. For instance, AIMNet \cite{liu2024attention} introduced an attention-guided embedding completion strategy, which dynamically extracted information from available views via cross-view attention propagation to reconstruct the representations of missing ones. SIP \cite{liu2024partial} established  a model aligned with the information bottleneck principle to learn cross-view shared representations, while MTD  \cite{liu2023masked} utilized a masked dual-channel decoupling framework to capture view-specific
information.

\section{Method}
\begin{figure*}[htbp] % 使用 figure* 环境使图片跨双栏
    \centering 
\includegraphics[width=\textwidth]{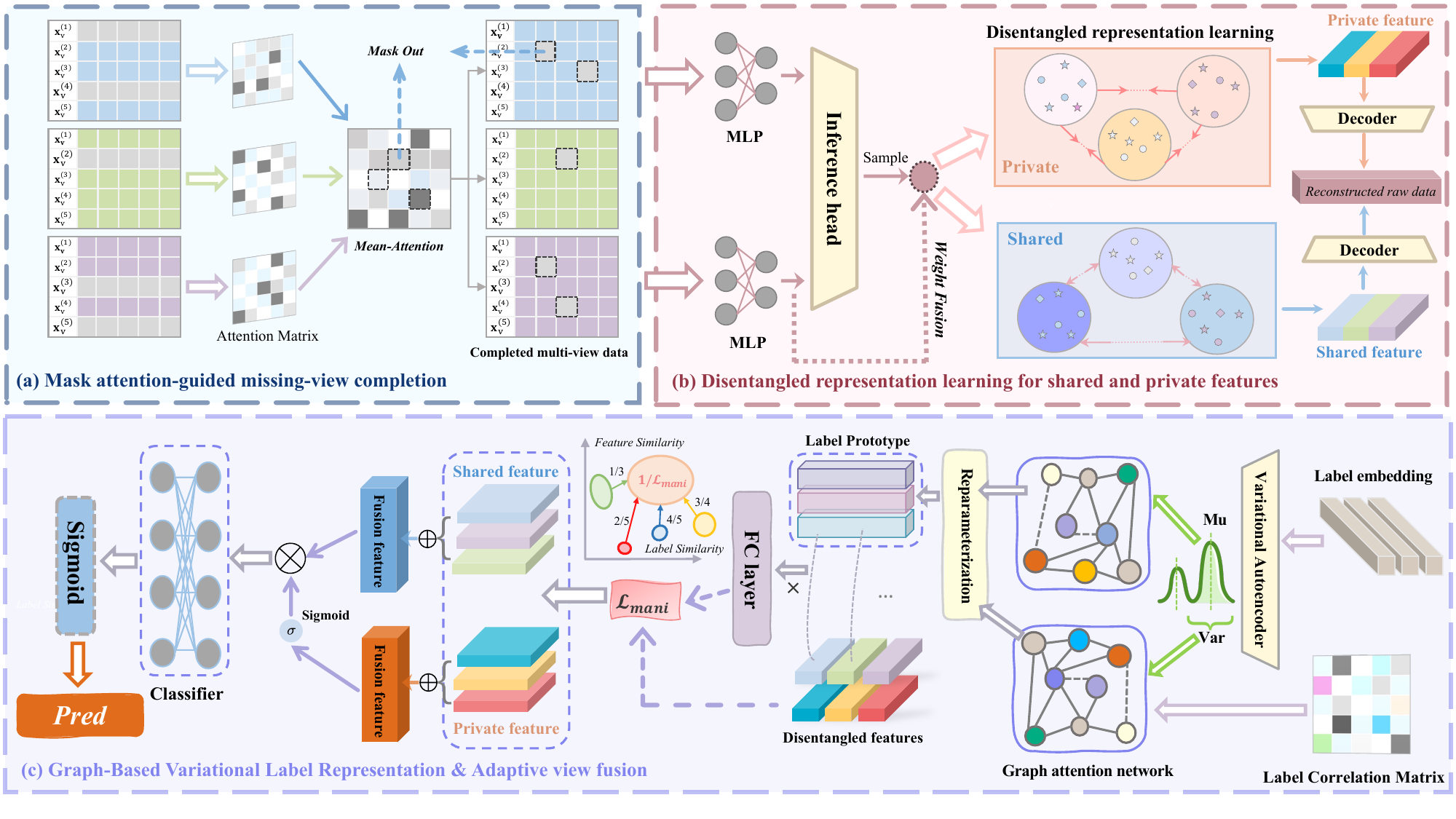}
    \caption{The main framework of our proposed ADRL. Different shapes signify different samples.}
    \label{fig:wide_image} % 用于在文本中交叉引用该图片，例如 \ref{fig:wide_image}
\end{figure*}

\subsection{Problem Formulation and Main Notation}
A given dataset comprising  $V$ views and $N$ samples can be  described in two equivalent forms: $\{\bm{X}^{(v)} \in \mathbb{R}^{N \times d_{v}}\}_{v=1}^{V}$ or $\{\{\bm{x}_{i}^{(v)}\}_{i=1}^{N}\}_{v=1}^{V}$, where $\bm{X}^{(v)}$ is the data matrix of the $v$-th view with feature dimension $d_v$, and $\bm{x}_{i}^{(v)}$ denotes its 
$i$-th sample. Let $\bm{Y} \in\{0,1\}^{N \times C}$ represent the binary label matrix, where 
$C$ is the total number of categories. $\bm{Y}_{i,j}=1$ indicates that the $i$-th sample is tagged as the $j$-th class, otherwise $\bm{Y}_{i,j}=0$. To  account for the  necessity to explicitly express data absence, we let $\bm{W} \in\{0,1\}^{N \times V}$ and $\bm{G} \in\{0,1\}^{N \times C}$ denote the 
 missing indicator for views and labels, respectively. Specifically, $\bm{W}_{i,j}=1$ means the $j$-th view of the $i$-th sample is available, otherwise $\bm{W}_{i,j}=0$. Similarly, $\bm{G}_{i,j}=1$ signifies that the corresponding label is known with certainty. Our goal is to develop a neural network model capable of learning from partially labeled incomplete multi-view data, while maintaining strong inference performance on unlabeled test cases.

\subsection{Mask Attention-Guided  Missing-View Completion}
In real-world scenarios, the absence of multi-view data is a common phenomenon, and the lack of diverse views  restricts  the capacity of deep neural models to capture high-level feature representations \cite{li2024deep}. Therefore, data augmentation through reconstruction techniques is crucial for ensuring the stability of the feature extraction process and improving the overall accuracy of the model \cite{jiang2024deep}. To mitigate the impact of information bottlenecks caused by incompleteness,  we propose an attention-driven approach for constructing structural topology graphs, which can effectively transmit the relational signals between instances and  facilitate the imputation of missing features. With respect to any view $v$,  the attention score for an instance pair is computed by 
\begin{equation}\label{eq01}
{A}_{i,j} ^{(v)} = e^(\frac{h(\bm{x}_{i} ^{(v)})h(\bm{x}_{j} ^{(v)T})}{\tau}),
\end{equation}
where $h$ is the $l_{2} \text {-}$norm normalization function used to eliminate heterogeneous scales across different dimensions and $\tau$ is the temperature parameter. Each row of the matrix $\bm{A}^{(v)}$ quantitatively measures the pairwise similarity between the corresponding sample with all other samples in the same view, which serves as a criterion for filtering out weakly correlated samples. Due to the referential consistency among different views of the same sample, the relationships established in the existing views can be extended to the missing views. Thus, the relational dependencies of missing samples pertaining to a particular view can be approximated by aggregating attention scores from the remaining available views, i.e., 
\begin{equation}
B_{i, j}^{(v)}=\frac{1}{\sum_{k=1, k \neq v} W_{i, k}W_{j, k}} \sum_{k=1, k \neq v}^{V} W_{i, k}W_{j, k} \tilde{A}_{i, j}^{(k)},
\end{equation}
where $\bm{B}^{(v)}$ is a sample affinity matrix independent of its original view and computed by integrating attention scores obtained from other valid views. $\tilde{\bm{A}}^{(v)}$ is a high-confidence version derived via threshold filtering, defined as $\tilde{\bm{A}}^{(v)}_{i,j}={\bm{A}}^{(v)}_{i,j}$ if ${\bm{A}}^{(v)}_{i,j}>\mathcal{T}$, otherwise $\tilde{\bm{A}}^{(v)}_{i,j}=0$. The aim of thresholding mechanism is to avoid the weakening of sparse yet high-attention  associations and  enhance the selection of strongly correlated samples. Then, a transferred graph $\bm{K}^{(v)} \in \mathbb{R}^{N \times N}$ is constructed to identify the available instances associated with the missing ones, where  $\bm{K}_{i, j}^{(v)}=1$ means that $\bm{W}_{j, v}=1$ and $\bm{B}_{i, j}^{(v)}$ is among the top-$k$ entries in the $i$-th row. By treating $\bm{K}$ as the adjacency matrix and $\bm{B}$ as the edge weights, the reconstructed data can be obtained through the process of message propagation:
\begin{equation}
\hat{\bm{x}}_{i}^{(v)}=\frac{\sum_{j=1}^{N} {K}_{i, j}^{(v)} {B}_{i, j}^{(v)}  \bm{x}_{j}^{(v)}}{\sum_{j=1}^{N}  {B}_{i, j}^{(v)}}.
\end{equation}
As $\hat{\bm{X}}$ serves as an approximate substitute for the missing instances,  it is integrated with the original available views to construct the final data matrix for downstream tasks:
\begin{equation} \label{eq04}
{\hat{\bm{Z}}}_{i,:}^{(v)}=\hat{\bm{X}}_{i,:}^{(v)}\left(1-{W}_{i, v}\right)+\bm{X}_{i,:}^{(v)} {W}_{i, v}.
\end{equation}

 multi-view data $\{\hat{\bm{Z}}^{(v)}\}_{v=1}^{V}$ often suffer from information redundancy and residual noise, which stem from  improperly collected features and imperfect alignment across modalities. Besides, 
 feature reconstruction inevitably performs below expectations on certain samples.  These undesired issues can obscure discriminative signals and impair the model’s ability to generalize.  Thus, we  adopt a random fragment masking strategy, a technique widely used in the image recognition tasks \cite{9879206}, to suppress the weak feature components and enhance the robustness of feature expressions. Specifically,  for each view $v$, we construct a binary masking matrix $\bm{M}^{(v)} \in\{0,1\}^{N \times d_{v}}$ matching the dimensionality of $\hat{\bm{Z}}^{(v)}$. To introduce structured corruption, we uniformly sample a position index $b_{i} \in[1, d_{v}-l]$ for each instance, and set a contiguous segment of length $l$ starting from $b_{i}$  in the $i$-th row of $\bm{M}^{(v)}$ to zero. The masked representation is then  derived as follows:
\begin{equation}\label{eq05}
\bm{Z}^{(v)}=\hat{\bm{Z}}^{(v)} \odot \bm{M}^{(v)},
\end{equation}
where $\odot$ denotes element-wise multiplication.

\subsection{Disentangled Representation Learning For Shared and Private Features}
A fundamental characteristic of multi-view data is the coexistence of both view-shared and view-specific information components \cite{LiLL25}. For example, in image recognition tasks with multi-angle photos of the same object, structural features like shape are consistent across views, while lighting or background details remain view-dependent.    Thus, our goal is to develop effective mechanisms for feature purification and achieve a complete separation of representation information. During the feature extraction process, we design dual channels to obtain the shared and private representations, denoted as $\{{\bm{Z}}^{(v)}_s \in \mathbb{R}^{N \times d}\}_{v=1}^{V}$ and $\{{\bm{Z}}^{(v)}_p \in \mathbb{R}^{N \times d}\}_{v=1}^{V}$, respectively. Both channels adopt the identical network architectures and perform forward propagation independently. The recovered multi-view data $\{{\bm{Z}}^{(v)}\}_{v=1}^{V}$ is firstly passed through a set of feature extractors to obtain the initial  representations $\{{\bm{Z}}^{(v)}_{s_{1}}\}_{v=1}^{V}$ and $\{{\bm{Z}}^{(v)}_{p_{1}}\}_{v=1}^{V}$. To effectively capture the feature interactions, it is crucial to model the statistical characteristics of multiple views. Inspired by the principle of variational inference \cite{45903}, the assumption that each view follows a Gaussian distribution facilitates the information integration of heterogeneous modalities and enables tractable posterior estimation. Subsequently, each view is processed by four additional encoders to approximate the underlying distribution of its shared and specific representations. These distributions are given by $p({\bm{Z}}^{(v)}_{s_{2}} \mid {\bm{Z}}^{(v)}):=\mathcal{N}(f_{\mu}^{v}({\bm{Z}}^{(v)}), f_{\sigma^{2}}^{v}({\bm{Z}}^{(v)}) \bm{I})$ and $p({\bm{Z}}^{(v)}_{p_{2}} \mid {\bm{Z}}^{(v)}):=\mathcal{N}(g_{\mu}^{v}({\bm{Z}}^{(v)}), g_{\sigma^{2}}^{v}({\bm{Z}}^{(v)}) \bm{I})$, where $f_{\mu}^{v}$ and $f_{\sigma^{2}}^{v}$ denote the mean and variance encoders for the shared information channel, while  $g_{\mu}^{v}$ and $g_{\sigma^{2}}^{v}$ for another channel. Then,  alternative representations $\{{\bm{Z}}^{(v)}_{s_{2}}\}_{v=1}^{V}$ and $\{{\bm{Z}}^{(v)}_{p_{2}}\}_{v=1}^{V}$ are obtained by sampling from the corresponding distributions. To regulate the discrepancy between the sampled representations and the initial feature embeddings while  improving the accuracy of distribution learning, the two types of representations are fused based on the derived statistical properties. By using the reciprocal of the variance across feature dimensions as fusion weights, stable features are assigned higher confidence. The final form of each dimension in the representation is formulated as
\begin{equation}\label{eq06}
\left\{\begin{array}{l}
{\bm{Z}}^{(v)}_s=\frac{1}{f_{\sigma^{2}}^{v}(\bm{Z}^{(v)})}\bm{Z}^{(v)}_{s_{2}}+(1-\frac{1}{f_{\sigma^{2}}^{v}(\bm{Z}^{(v)})})\bm{Z}^{(v)}_{s_{1}} \\
{\bm{Z}}^{(v)}_p=\frac{1}{g_{\sigma^{2}}^{v}(\bm{Z}^{(v)})}\bm{Z}^{(v)}_{p_{2}}+(1-\frac{1}{g_{\sigma^{2}}^{v}(\bm{Z}^{(v)})})\bm{Z}^{(v)}_{p_{1}}.
\end{array}\right.
\end{equation}

The core objective of feature disentanglement is to maximize the information interaction among shared representations to highlight consistency, while suppressing the overlap among private features to emphasize complementarity \cite{bao2021disentangled}. To obtain consensus representations, we extract cross-view invariant features by maximizing mutual information between heterogeneous perspectives: 
\begin{equation} 
\begin{aligned}
\label{eq:v1}
    \max \frac{1} {V}  \sum_{v=1}^{V}  \left(I(\bm{Z}^{(v)} ;   {\bm{Z}}^{(v)}_s) + \frac{\gamma}{(V-1)}  \sum_{ u\neq v} I({\bm{Z}}^{(u)};{\bm{Z}}^{(v)}_s) \right),
\end{aligned}
\end{equation}
where $\lambda$ is the balance coefficient and  maximizing $I(\bm{Z}^{(v)}; {\bm{Z}}^{(v)}_s)$ is crucial to ensure that the representations remain aligned with the fundamental semantics of the raw modalities. By optimizing the second term,  we explicitly encourage the shared representations to retain maximal mutual information across multiple views, thereby ensuring global cross-view consistency and semantic alignment. For private features, we design the following target to restrict ${\bm{Z}}^{(v)}_p$ to  contain only view-specific information:
\begin{equation} \label{eq:v2}
    \max \frac{1}{V} \sum_{v=1}^V \left(I(\bm{Z}^{(v)};   {\bm{Z}}^{(v)}_p) - \frac{\beta}{(V-1)} \sum_{u \neq v}I({\bm{Z}}^{(u)};{\bm{Z}}^{(v)}_p) \right),
\end{equation}
where the term $- \beta \sum_{u \neq v} I({\bm{Z}}^{(u)}_p;{\bm{Z}}^{(v)}_p)$ actively penalizes cross-view information overlap and preserves the distinct semantics of each view.

Since the exact computation of mutual information is intractable in high-dimensional spaces, we derive its bound that admits reliable estimation, which in turn supports the optimization of Eqs. (\ref{eq:v1}) and (\ref{eq:v2}). Building upon the fundamental properties of entropy and mutual information, we derive the following decomposition:
\begin{align}\label{mutual}
I(\bm{Z}^{(u)};{\bm{Z}}^{(v)}_s) &= H({\bm{Z}}^{(v)}_s) - H({\bm{Z}}^{(v)}_s \mid \bm{Z}^{(u)}) \notag \\
&= \underbrace{\left[H({\bm{Z}}^{(v)}_s) - H({\bm{Z}}^{(v)}_s \mid {\bm{Z}}^{(u)}_s)\right]}_{I({\bm{Z}}^{(v)}_s; {\bm{Z}}^{(u)}_s)} \notag \\
&\quad + \underbrace{\left[H({\bm{Z}}^{(v)}_s \mid {\bm{Z}}^{(u)}_s) - H({\bm{Z}}^{(v)}_s \mid {\bm{Z}}^{(u)}_s \bm{Z}^{(u)})\right]}_{I({\bm{Z}}^{(v)}_s; {\bm{Z}}^{(u)}_s \mid \bm{Z}^{(u)})} \notag \\
&\quad - \underbrace{\left[H({\bm{Z}}^{(v)}_s \mid \bm{Z}^{(u)}) - H({\bm{Z}}^{(v)}_s \mid {\bm{Z}}^{(u)}_s \bm{Z}^{(u)})\right]}_{I({\bm{Z}}^{(v)}_s; \bm{Z}^{(u)} \mid {\bm{Z}}^{(u)}_s)} \notag \\
&\geq I({\bm{Z}}^{(v)}_s; {\bm{Z}}^{(u)}_s) + I({\bm{Z}}^{(v)}_s; \bm{Z}^{(u)} \mid {\bm{Z}}^{(u)}_s) \notag \\
&\geq I({\bm{Z}}^{(v)}_s; {\bm{Z}}^{(u)}_s).
\end{align}
Inspired by \cite{bao2021disentangled}, we design a neural estimator based on the contrastive learning paradigm to approximate $I({\bm{Z}}^{(v)}_s; {\bm{Z}}^{(u)}_s)$. In the contrastive framework, positive pairs originate from the same instance, whereas negatives are formed by randomly matching samples. Then, we adopt the Jensen-Shannon divergence (JSD) lower bound of mutual information for approximation, which can be expressed as
\begin{equation}\label{eqqsd}
\begin{aligned}
\hat{{I}}_{\mathrm{JSD}}(\bm{X}^{(1)}, \bm{X}^{(2)}) &= \frac{1}{N} \sum_{i=1}^{N} \left[
-\log \left(1+e^{-T\left(\bm{x}^{(1)}_{i} \oplus \bm{x}^{(2)}_{i}\right)}\right) \right. \\
&\left. \quad -\log \left(1+e^{T\left(\bm{x}^{(1)}_{{\pi(i)}} \oplus \bm{x}^{(2)}_{i}\right)}\right)
\right],
\end{aligned}
\end{equation}
where $\pi$ denotes a random permutation over indices $\{1,2, \ldots, N\}$ and $\oplus$ is the concatenate operation. $T$ is a scoring function implemented via a multilayer perceptron. In Eq. (\ref{eqqsd}), as the mutual information between two samples increases, the score produced by $T$ becomes smaller. Through the contrastive learning, the statistical dependencies between two views can be captured, which provides an indirect estimation of mutual information.

Regarding $I({\bm{Z}}^{(u)};{\bm{Z}}^{(v)}_p)$ in Eq. (\ref{eq:v2}), we derive its variational upper bound through the following transformation:
\begin{equation}
\begin{aligned}\label{eq1010}
I(\bm{Z}^{(u)}, {\bm{Z}}^{(v)}_p) 
&= KL[p({\bm{Z}}^{(v)}_p|\bm{Z}^{(u)})p(\bm{Z}^{(u)})\|p({\bm{Z}}^{(v)}_p)p(\bm{Z}^{(u)})] \\
&=KL[p({\bm{Z}}^{(v)}_p;\bm{Z}^{(u)})\|r({\bm{Z}}^{(v)}_p)p(\bm{Z}^{(u)})]
\\
&
-  KL[r({\bm{Z}}^{(v)}_p)\|p({\bm{Z}}^{(v)}_p)]\\
&
\leq \mathbb{E}_{p(\bm{Z}^{(u)},{\bm{Z}}^{(v)}_p)}\log\frac{p({\bm{Z}}^{(v)}_p|\bm{Z}^{(u)})}{r({\bm{Z}}^{(v)}_p)} \\
&= \mathbb{E}_{p(\bm{Z}^{(u)},{\bm{Z}}^{(v)}_p)}\log \frac{p({\bm{Z}}^{(v)}_p|\bm{Z}^{(u)})}{p({\bm{Z}}^{(u)}_p|\bm{Z}^{(u)})} 
\\
&
+
\mathbb{E}_{p(\bm{Z}^{(u)},{\bm{Z}}^{(v)}_p)}\log \frac{p({\bm{Z}}^{(u)}_p|\bm{Z}^{(u)})}{r({\bm{Z}}^{(v)}_p)}.
\end{aligned}
\end{equation}
To ensure efficient and stable computation, we employ a standard normal distribution $r({\bm{Z}}^{(v)}_p)$ as a proxy for the marginal distribution $p({\bm{Z}}^{(v)}_p)$ \cite{lian2023online}. Since the two terms of the upper bound derived in Eq. (\ref{eq1010}) share a consistent optimization direction, i.e., maximizing the discrepancy between ${\bm{Z}}^{(v)}_p$ and  ${\bm{Z}}^{(u)}_p$, we simplify the upper bound into a tractable surrogate $\mathbb{E}_{p(\bm{Z}^{(u)},{\bm{Z}}^{(v)}_p)}\log \frac{p({\bm{Z}}^{(u)}_p|\bm{Z}^{(u)})}{r({\bm{Z}}^{(v)}_p)}$. Therefore, the loss used for feature disentanglement is as follows:
\begin{equation}
\begin{aligned}
\mathcal{L}_{\text{dis}} &=  \sum_{v=1}^{V}   \sum_{u=1, u\neq v}^{V}\left(- \frac{\gamma}{V(V-1)} \hat{{I}}_{\mathrm{JSD}}({\bm{Z}}^{(u)};{\bm{Z}}^{(v)}_s)  \right. \\ & \left. +\frac{\beta}{V(V-1)}\mathbb{E}_{p(\bm{Z}^{(u)},{\bm{Z}}^{(v)}_p)}\log \frac{p({\bm{Z}}^{(u)}_p|\bm{Z}^{(u)})}{r({\bm{Z}}^{(v)}_p)} \right),
\end{aligned}
\end{equation}
where $\gamma$ and $\beta$ are fixed to 0.01 to ensure numerical stability of loss computation. The first terms in Eqs. (\ref{eq:v1}) and (\ref{eq:v2}) are typically bounded by the reconstruction losses\cite{huang2023generalized}, where the obtained representations are decoded via stochastic decoders to  faithfully preserve the original view:
\begin{equation}
\begin{aligned}
\mathcal{L}_{\text{re}} &=\frac{1}{V} \sum_{v=1}^{V} \left(\mathbb{E}_{p\left(\bm{Z}^{(v)};   {\bm{Z}}^{(v)}_s\right)}\left[\log q(\bm{Z}^{(v)} | {\bm{Z}}^{(v)}_s)\right]
\right.\\ & \left. +\mathbb{E}_{p\left(\bm{Z}^{(v)};   {\bm{Z}}^{(v)}_p\right)}\left[\log q(\bm{Z}^{(v)} | {\bm{Z}}^{(v)}_p)\right]\right).
\end{aligned}
\end{equation}

\subsection{Graph-Based Variational Label Representation}
In this subsection, we aim to construct label prototypes while ensuring their correspondence with the correlation patterns inherent in the ground-truth labels. In order to explicitly model the expression of label semantic and enhance the robustness of multi-label learning, we employ stochastic encoders to fit the underlying distribution $\mathcal{N}(\mu_{i}, \sigma_{i}^{2} \bm{I})$ of each label prototype, where $\mu_{i}$ and $\sigma_{i}^{2}$  are the d-dimensional mean and variance outputs generated by the encoders $h_{\mu}(\bm{b}_i)$ and $h_{\sigma^{2}}(\bm{b}_i)$. $\bm{b}_i  \in \mathbb{R}^{C}$ is a learnable tensor associated with the $i$-th category, initialized as an one-hot vector with the $i$-th entry set to 1. This tensor is jointly optimized with other network parameters during training.  Besides, we leverage the co-occurrence frequencies of label pairs in the training data to quantitatively represent label correlations. The  correlation matrix $\bm{Q} \in \mathbb{R}^{C \times C}$ is computed as
\begin{equation} \label{eq144}
\bm{Q}_{i,j}=\frac{\sum_{k=1}^{N} {Y}_{k, i}{G}_{k, i} {Y}_{k, j}{G}_{k, j}}{\sum_{k=1}^{N}{Y}_{k, i}{G}_{k, i}}.
\end{equation}
Next, the  correlation information is transmitted through the graph attention network (GAT)  via $\bm{Q}$ as a bridge to refine the distribution parameters. For each head, we compute the attention coefficient as below:
\begin{equation}
\alpha_{ij}=\frac{exp({\sigma_L(\bm{a}\cdot([\bar{\bm{W}}z_i\oplus \bar{\bm{W}}z_j])))}}{\sum_{k\in\mathcal{N}_i}exp({\sigma_L(\bm{a}\cdot([\bar{\bm{W}}z_i\oplus \bar{\bm{W}}z_k])))}},
\end{equation}
where  $\oplus$  is the concatenate operation and   $z$ denotes the distribution parameters, that is, the set of parameters including the mean and variance are all subject to computing $\alpha_{ij}$. $\bm{a} \in  \mathbb{R}^{2d}$ and $\bar{\bm{W}} \in \mathbb{R}^{d\times d} $ are the learnable weights and $\sigma_L$  is the LeakyReLU activation function. Specifically,  $k \in \mathcal{N}_i$ indicates that we focus solely on related labels, i.e., $\bm{Q}_{i,k} > 0$. Then, we can obtain the enhanced label distribution parameters through information aggregation:
\begin{equation}
o_i^{\prime}=\sigma_L\left(\frac{1}{K}\sum_{k=1}^K\sum_{j\in\mathcal{N}_i}\alpha_{ij}^kW^kz_j\right),
\end{equation}
where $K$ represents the number of attention heads, with  $\alpha_{ij}^k$ and $W^k$ denoting the attention coefficient and weight parameter of the $k$-th head, respectively. $o_i^{\prime}$ is the final distribution parameters $\mu'_i$ or ${\sigma'_i}^2$, which  encode the semantic relationships among multiple labels. Formally, for each label $i$ with post-GAT parameters $(\mu'_i, {\sigma'_i}^2)$, we sample its embedding feature as
\begin{equation}\label{eq017}
\bm{l}_i = \mu'_i + \epsilon_i \odot \sigma'_i, \quad \epsilon_i \sim \mathcal{N}(0,1).
\end{equation}

\subsection{Adaptive View Fusion and  Multi-Label Classification}
Upon obtaining the shared and specific representations along with the class embeddings, the subsequent task is to architect a method for view fusion and facilitate the interaction between view representations and label prototypes to derive predictive outcomes. In classification problems involving multiple categories of features and labels,  the manifold assumption \cite{li2025semi} that samples with similar features are likely to share similar labels can provide crucial guidance for prediction and enhance the  classification discriminability. We begin by performing element-wise multiplication between the label prototypes modulated by the Sigmoid function and view representations, which yields the interaction-level representations:
\begin{equation}\label{eq144}
\hat{\bm{P}}_{i}^{(v)}=\left[\sigma_{S}\left(\bm{l}_{1}\right) \odot \bm{x}_{i}^{(v)} ; \sigma_{S}\left(\bm{l}_{2}\right) \odot \bm{x}_{i}^{(v)} ; \ldots ; \sigma_{S}\left(\bm{l}_{C}\right) \odot \bm{x}_{i}^{(v)}\right].
\end{equation}
According to Eq. (\ref{eq144}), we can obtain the label-specific features $\{{\bm{Z}}^{(v)}_s\rightarrow \hat{\bm{U}}^{(v)} \in \mathbb{R}^{N \times C \times d}\}_{v=1}^{V}$ and $\{{\bm{Z}}^{(v)}_p
\rightarrow \hat{\bm{V}}^{(v)} \in \mathbb{R}^{N \times C \times d}\}_{v=1}^{V}$. 
By capturing the distinct dependencies of label prototypes on individual views, these label-specific features create opportunities for adaptive view selection, which allows for  the identification of the most discriminative views for each category. Then, processing through the linear classifiers, view-specific predictions ${\bm{U}}^{(v)} \in \mathbb{R}^{N \times C}$ and ${\bm{V}}^{(v)} \in \mathbb{R}^{N \times C }$ are generated. To ensure consistency in sample relationships between the feature and label space, the learned feature representations should conform to the correlation structure exhibited by samples under multiple labels. The following manifold loss is computed to quantify this alignment:
\begin{equation} \label{eq019}
\begin{aligned}
\mathcal{L}_{g c}^{(v)} & =-\frac{1}{N(N-1)}\sum_{i=1}^{N} \sum_{j \neq i}^{N}({T}_{i,j}^{(v)} \log {S}_{i,j}^{(v)}  \\
&+(1-{T}_{i,j}^{(v)}) \log (1-{S}_{i,j}^{(v)})),
\end{aligned}
\end{equation}
where $\bm{S}_{i,j}^{(v)}=\frac{1 +\bm{x}_{i}^{(v)} {(\bm{x}_{j}^{(v)})}^T}{2}$ represents the feature similarity. Besides,  $\bm{T}_{i,j}^{(v)}={\bm{P}_i}^{(v)} {(\bm{P}^{(v)}_j)}^T$, where ${\bm{P}}^{(v)}$ corresponds to the  $L_2$-normalized form of ${\bm{U}}^{(v)}$ or  ${\bm{V}}^{(v)}$. A higher value of the loss $\mathcal{L}_{g c}^{(v)}$ signifies weaker structural consistency between the corresponding feature and multiple labels, indicating a low-quality and unreliable representation. Thus, such feature should be assigned smaller  weights  in the fusion process. 
The reciprocal of the loss $\mathcal{L}{gc}^{(v)}$  is then utilized as the fusion coefficient, which enables the derivation of the  integrated shared representation $\bm{Z}_s$
 and the specific representation $\bm{Z}_p$
 via weighted summation. Moreover, the losses $\mathcal{L}_{g c}^{(v)}$  computed from all shared and private representations are averaged and scaled by  a factor of 0.05 to yield the training loss $\mathcal{L}_{g c}$.
 
 \begin{figure*}[t!] % figure* 表示图片跨双栏，htbp 是浮动参数
    \centering % 整体居中

    % 第一行图片
    \subfloat[Corel5k\label{fig:image1}]{%
        \includegraphics[width=0.31\textwidth]{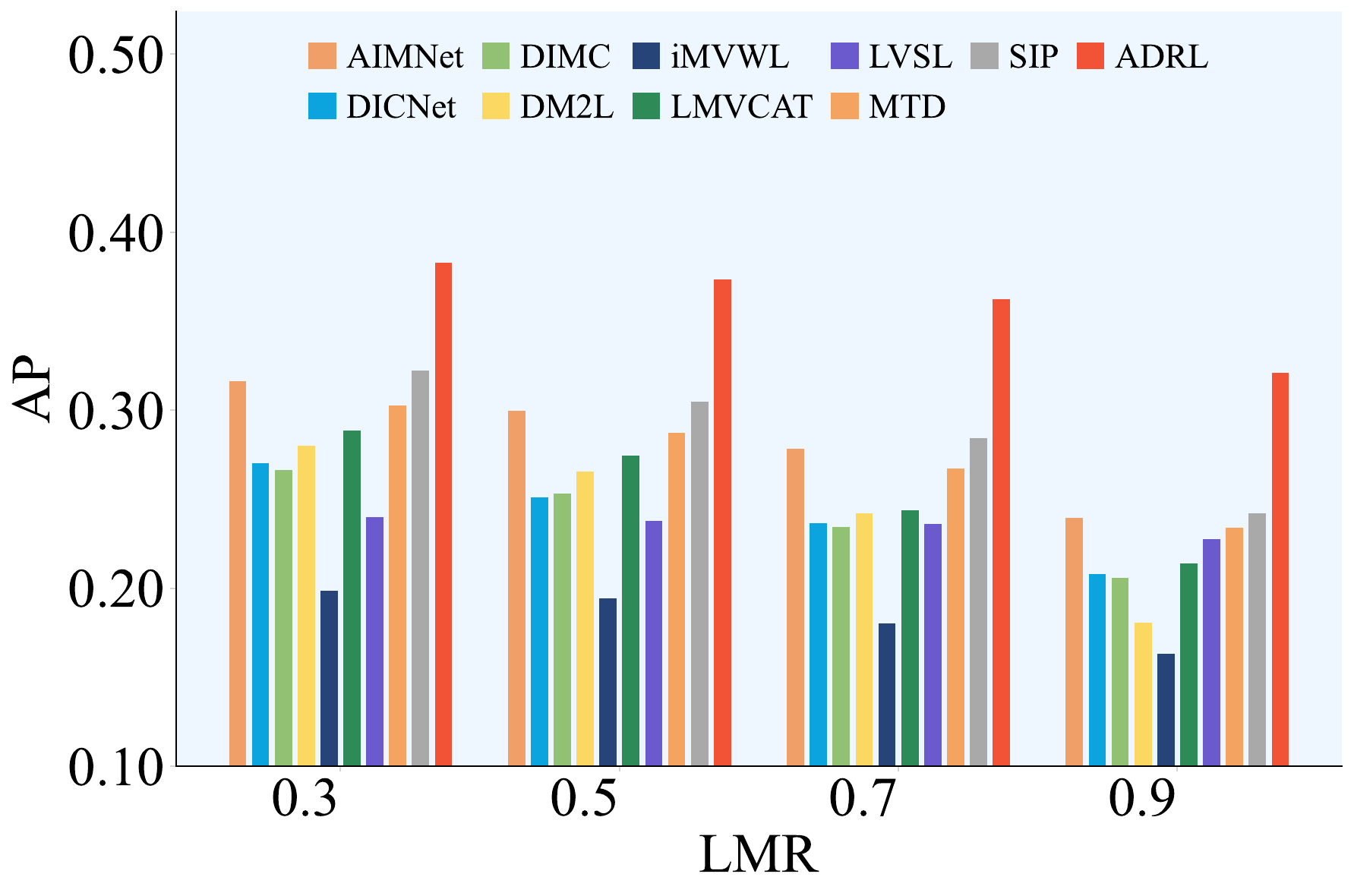}% 宽度略微调整以适应间距，可以按需修改
    }% 注意这里的百分号，可以避免不必要的空格，也可以省略，\hfill 会处理间距
    \hfill % 水平填充，使图片之间有间隔
    \subfloat[ESPGame\label{fig:image2}]{%
        \includegraphics[width=0.31\textwidth]{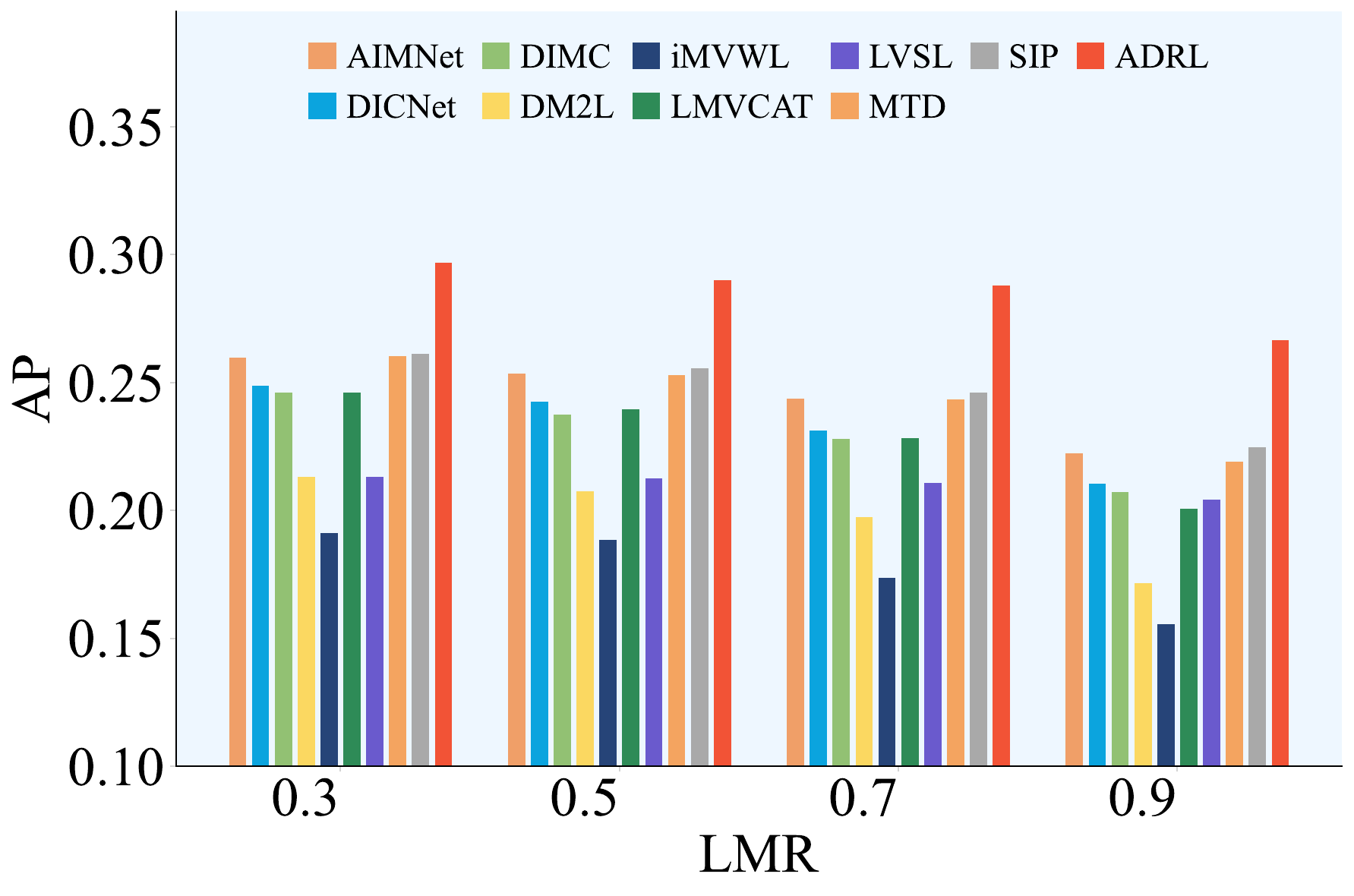}%
    }%
    \hfill % 水平填充
    \subfloat[IAPRTC12\label{fig:image3}]{%
        \includegraphics[width=0.31\textwidth]{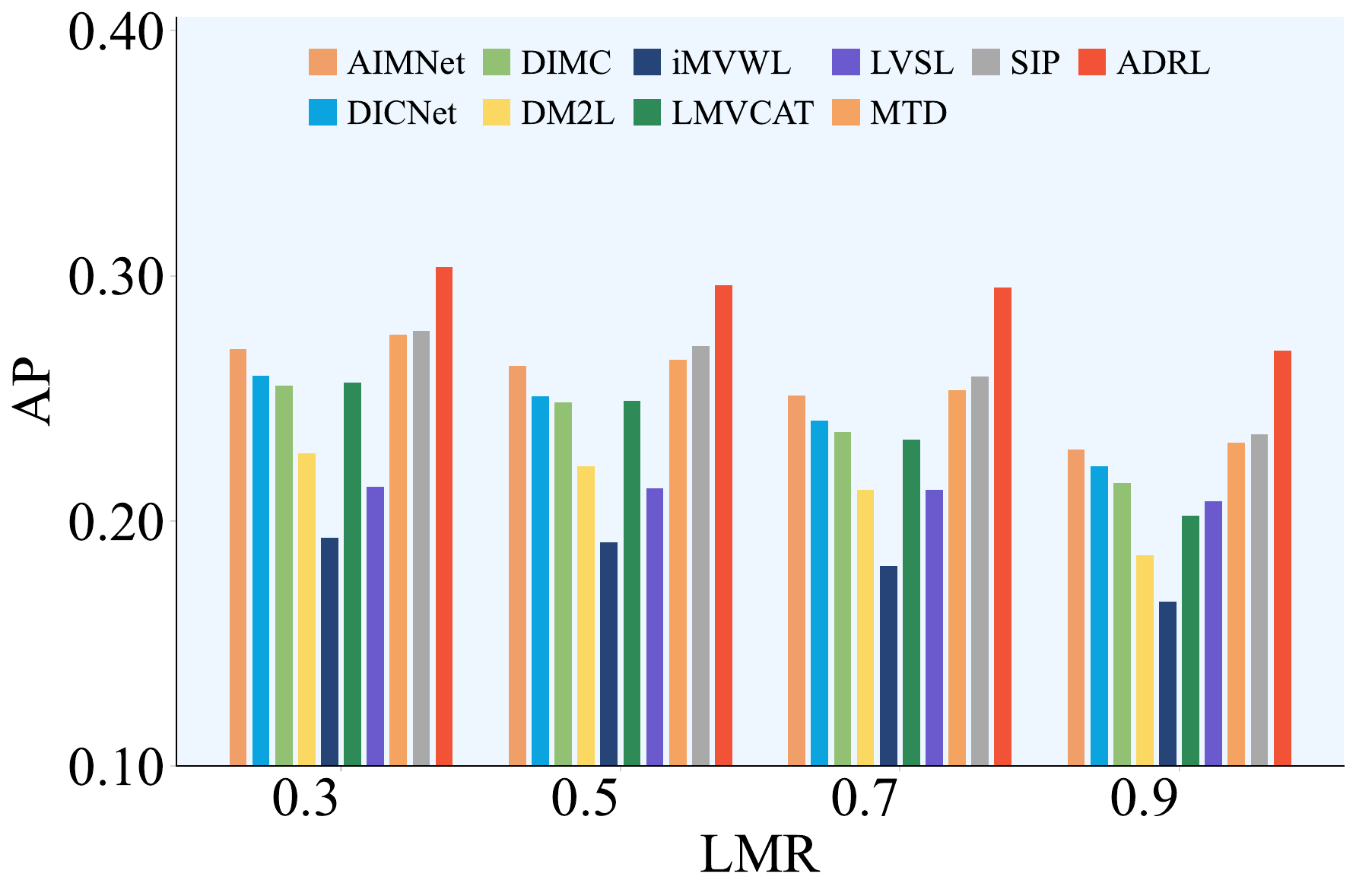}%
    }%
    \\ % 换行，开始新的一行子图

    % 第二行图片
    \subfloat[Mirflickr\label{fig:image4}]{%
        \includegraphics[width=0.31\textwidth]{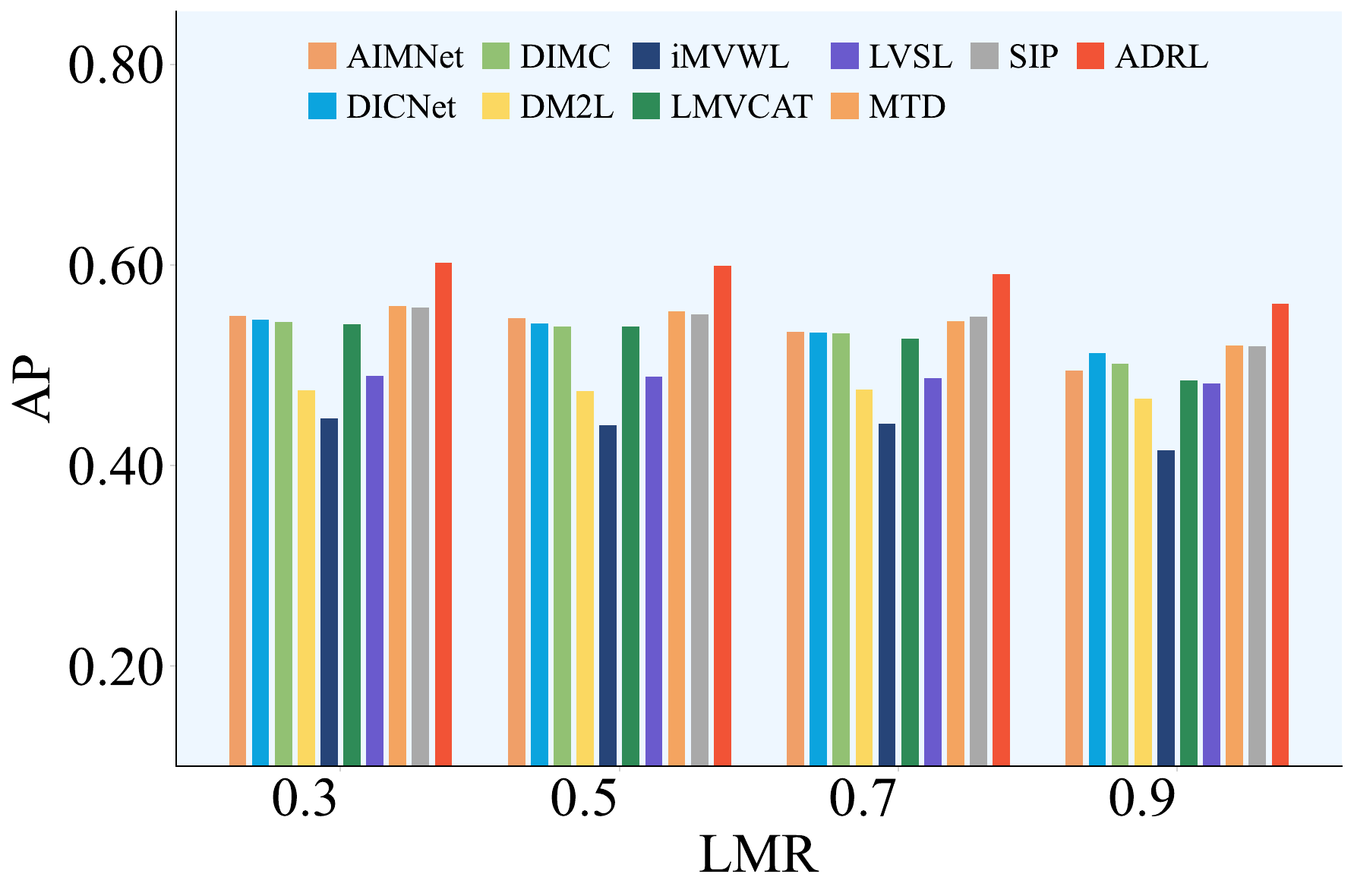}%
    }%
    \hfill % 水平填充
    \subfloat[Object\label{fig:image5}]{%
        \includegraphics[width=0.31\textwidth]{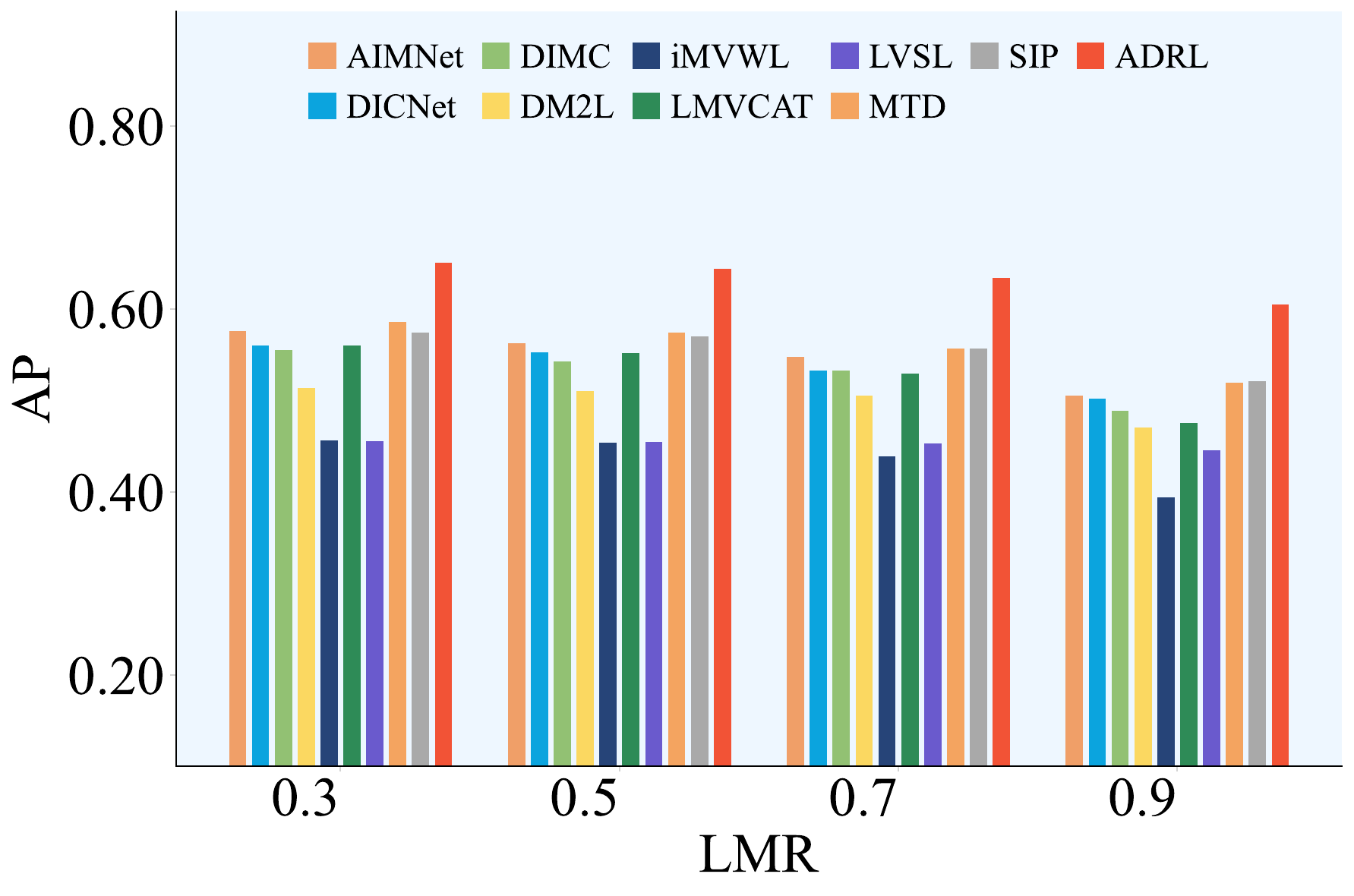}%
    }%
    \hfill % 水平填充
    \subfloat[Pascal07\label{fig:image6}]{%
        \includegraphics[width=0.31\textwidth]{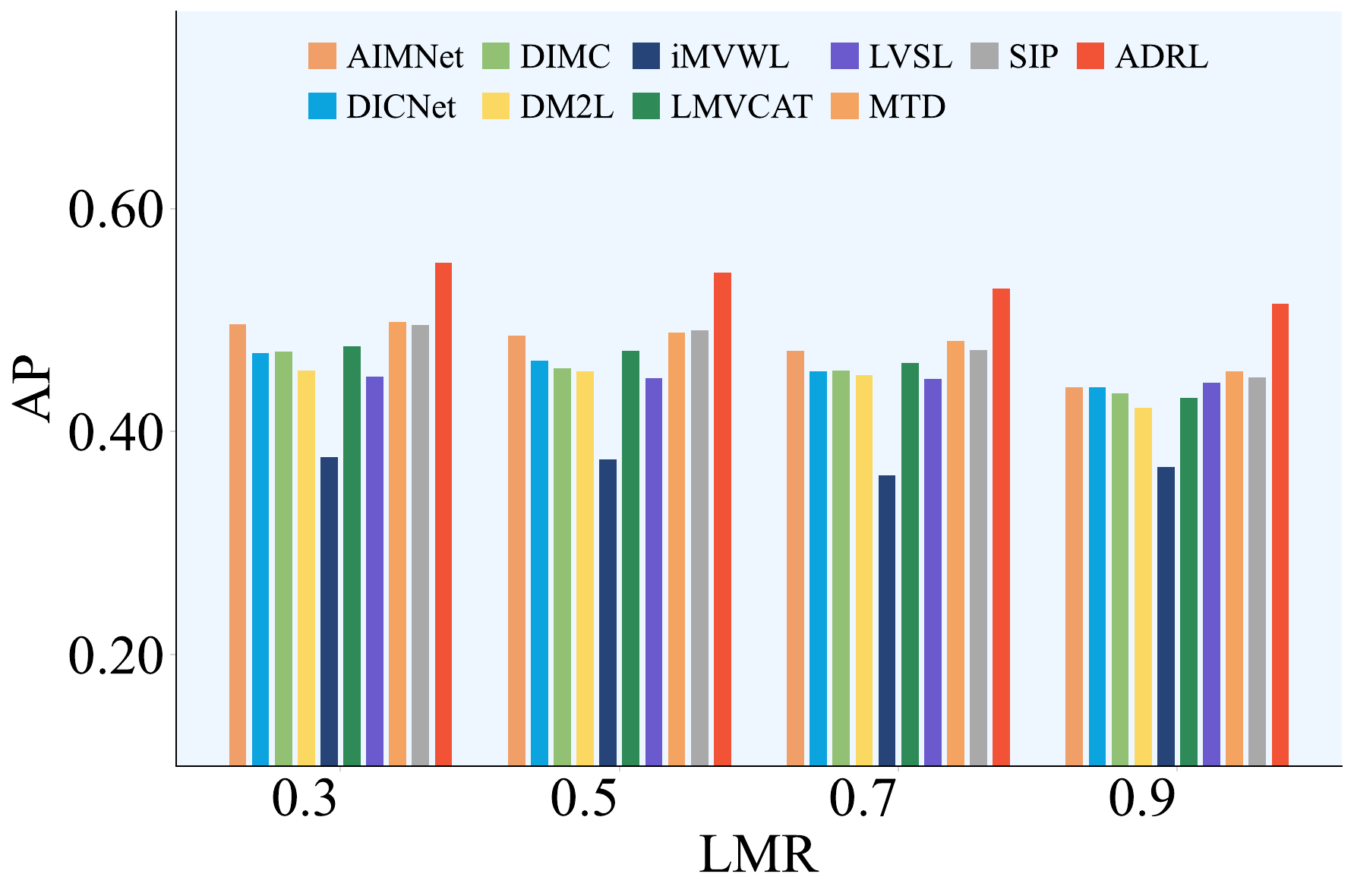}%
    }%
    \caption{Performance under different LMR with FMR fixed at 0.9.}
    \label{figII}
\end{figure*}
 
 To synthesize representation information from different aspects, we enhance  shared representation through the incorporation of view-specific information:
\begin{equation}\label{Eq020}
\bm{Z}=\sigma_S ({\bm{Z}_{p}}) \odot {\bm{Z}_{s}}, 
\end{equation}
where $\bm{Z} \in \mathbb{R}^{N \times d}$ is the final fused representation. The next step involves deriving the final prediction $\bm{P}$ in the same manner as obtaining ${\bm{U}}^{(v)}$ and ${\bm{V}}^{(v)}$. Then, we employ the weighted cross-entropy loss to mitigate the impact of missing labels, while  improving the model's classification capability:
\begin{align}
\mathcal{L}_{\text{mce}} = &-\frac{1}{\sum_{i,j}\bm{G}_{i,j}} \sum_{i=1}^N \sum_{j=1}^C \left( \bm{Y}_{i,j} \log({\bm{P}}_{i,j}) \right. \nonumber \\
&+ \left. (1 - \bm{Y}_{i,j}) \log(1 - {\bm{P}}_{i,j}) \right) \bm{G}_{i,j}.
\end{align}
Additionally, for each generated pseudo-label ${\bm{U}}^{(v)}$ and ${\bm{V}}^{(v)}$, we also compute $\mathcal{L}_{\text{mce}}$ to enhance their credibility, and average the results to obtain the loss $\mathcal{L}_{\text{pmce}}$.

By introducing the parameters $\alpha$, $\lambda_{1}$ and $\lambda_{2}$ to balance the loss effects, the total training loss can be expressed as
\begin{equation} \label{Eq022}
\mathcal{L} = \alpha \mathcal{L}_{\text{mce}} + \lambda_1 \mathcal{L}_{\text{re}} + \lambda_2 \mathcal{L}_{\text{pmse}} +\mathcal{L}_{\text{gc}} + \mathcal{L}_{\text{dis}}.
\end{equation}

\section{Experiments}
% In this section, we will provide a detailed introduction to the dataset settings, comparison methods, and analyze the experimental results.
\begin{figure*}[t!] % figure* 表示图片跨双栏，htbp 是浮动参数
    \centering % 整体居中

    % 第一行图片
    \subfloat[Corel5k\label{fig:image1}]{%
        \includegraphics[width=0.31\textwidth]{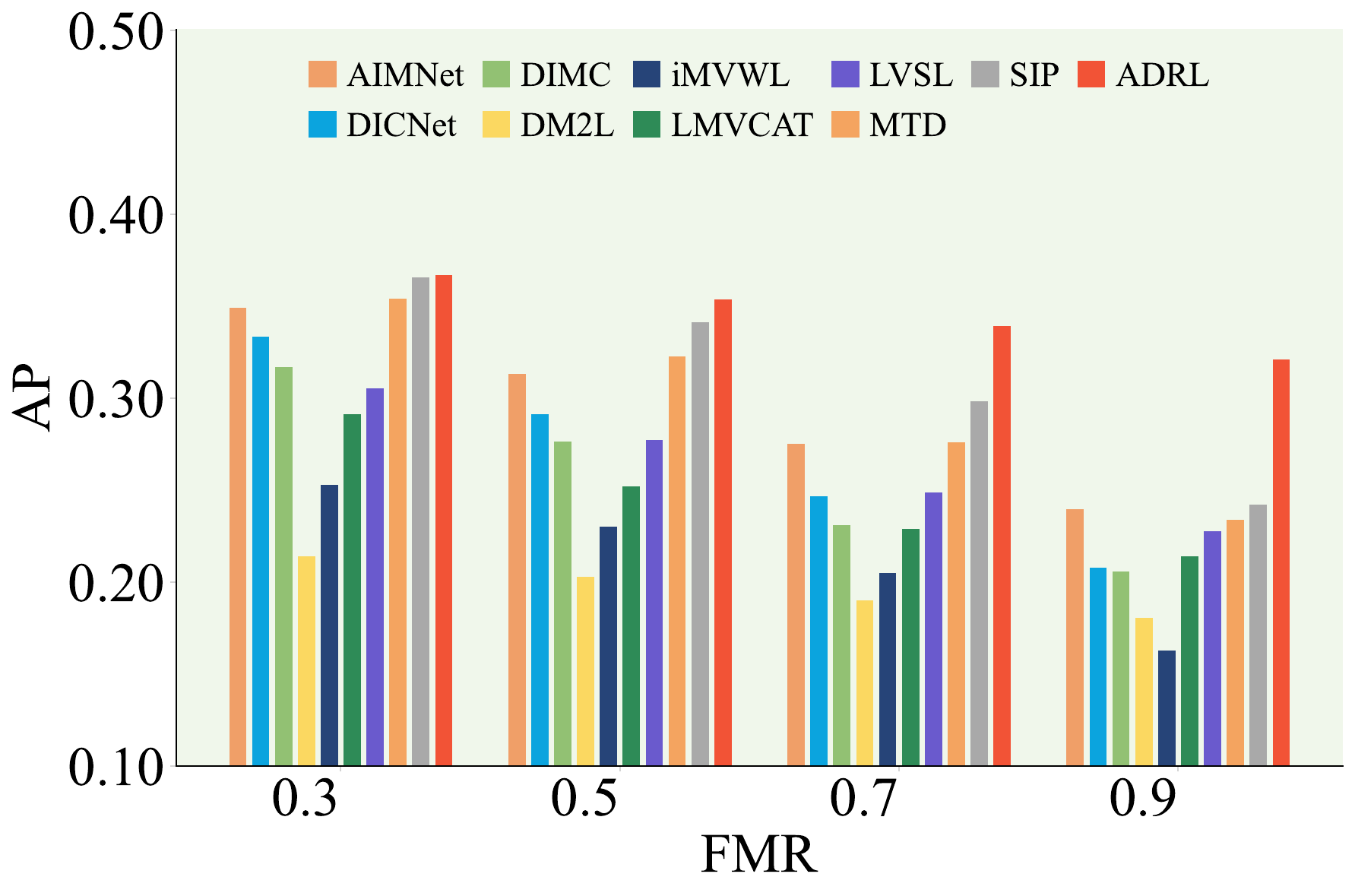}% 宽度略微调整以适应间距，可以按需修改
    }% 注意这里的百分号，可以避免不必要的空格，也可以省略，\hfill 会处理间距
    \hfill % 水平填充，使图片之间有间隔
    \subfloat[ESPGame\label{fig:image2}]{%
        \includegraphics[width=0.31\textwidth]{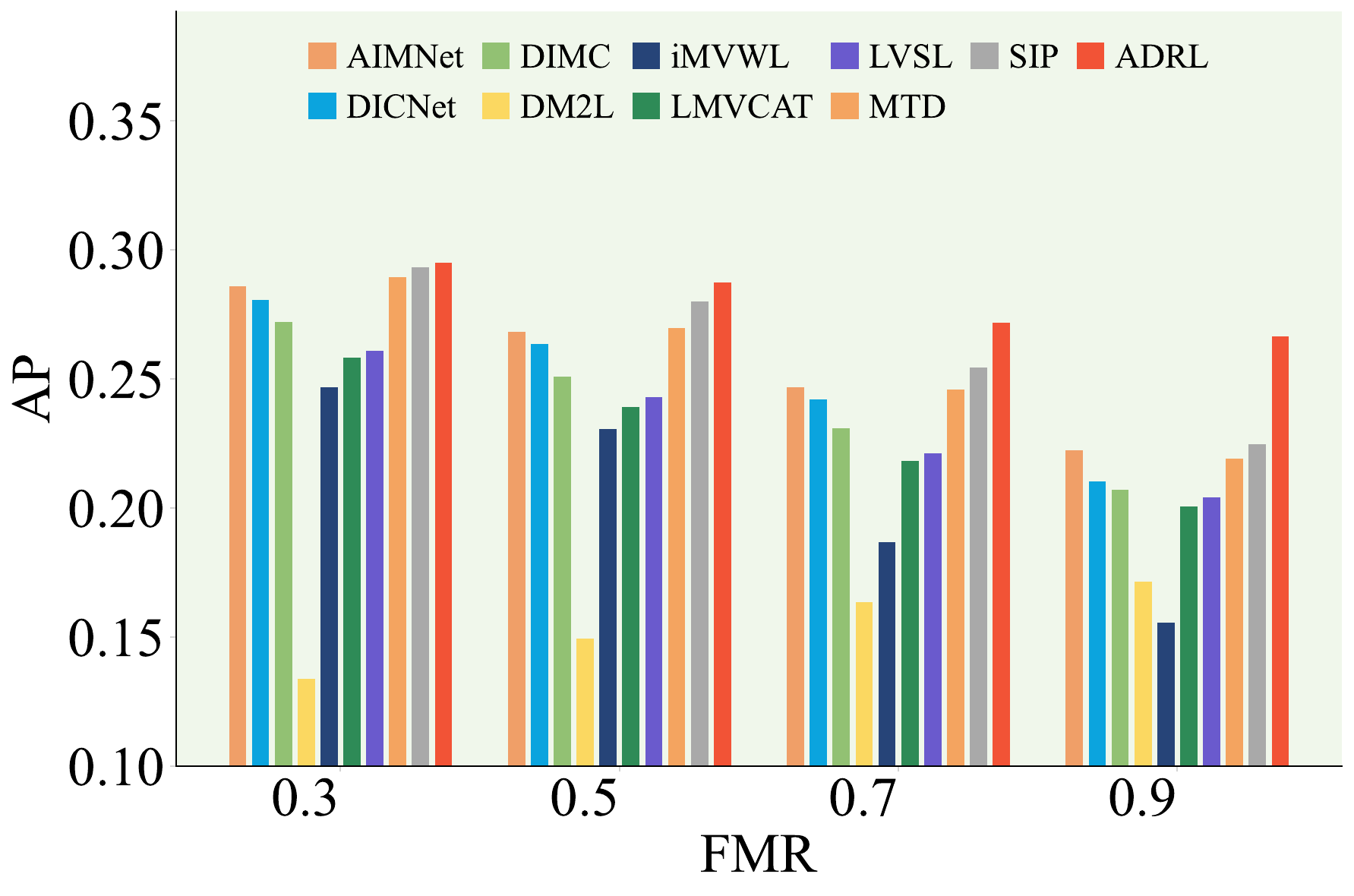}%
    }%
    \hfill % 水平填充
    \subfloat[IAPRTC12\label{fig:image3}]{%
        \includegraphics[width=0.31\textwidth]{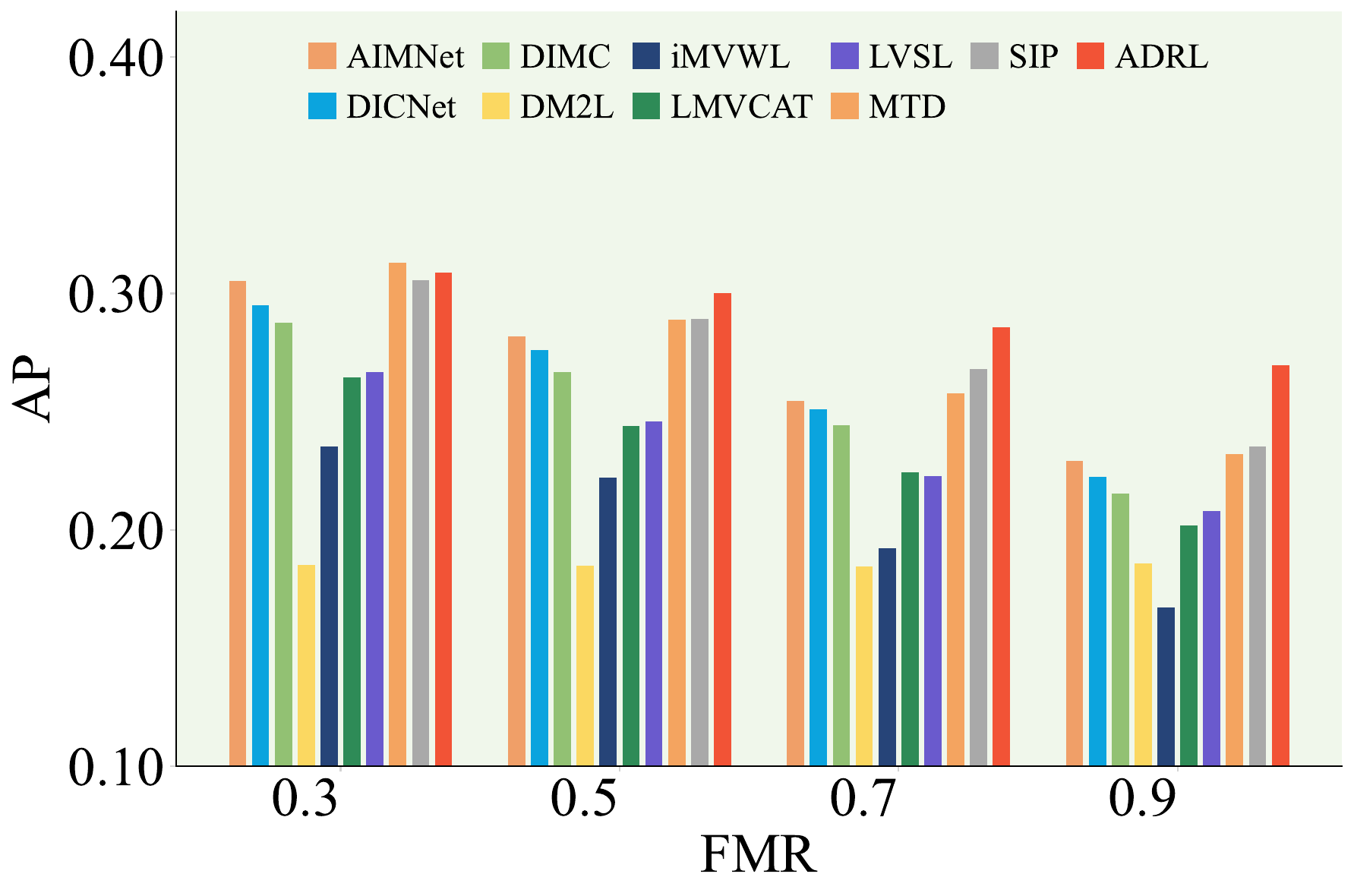}%
    }%
    \\ % 换行，开始新的一行子图

    % 第二行图片
    \subfloat[Mirflickr\label{fig:image4}]{%
        \includegraphics[width=0.31\textwidth]{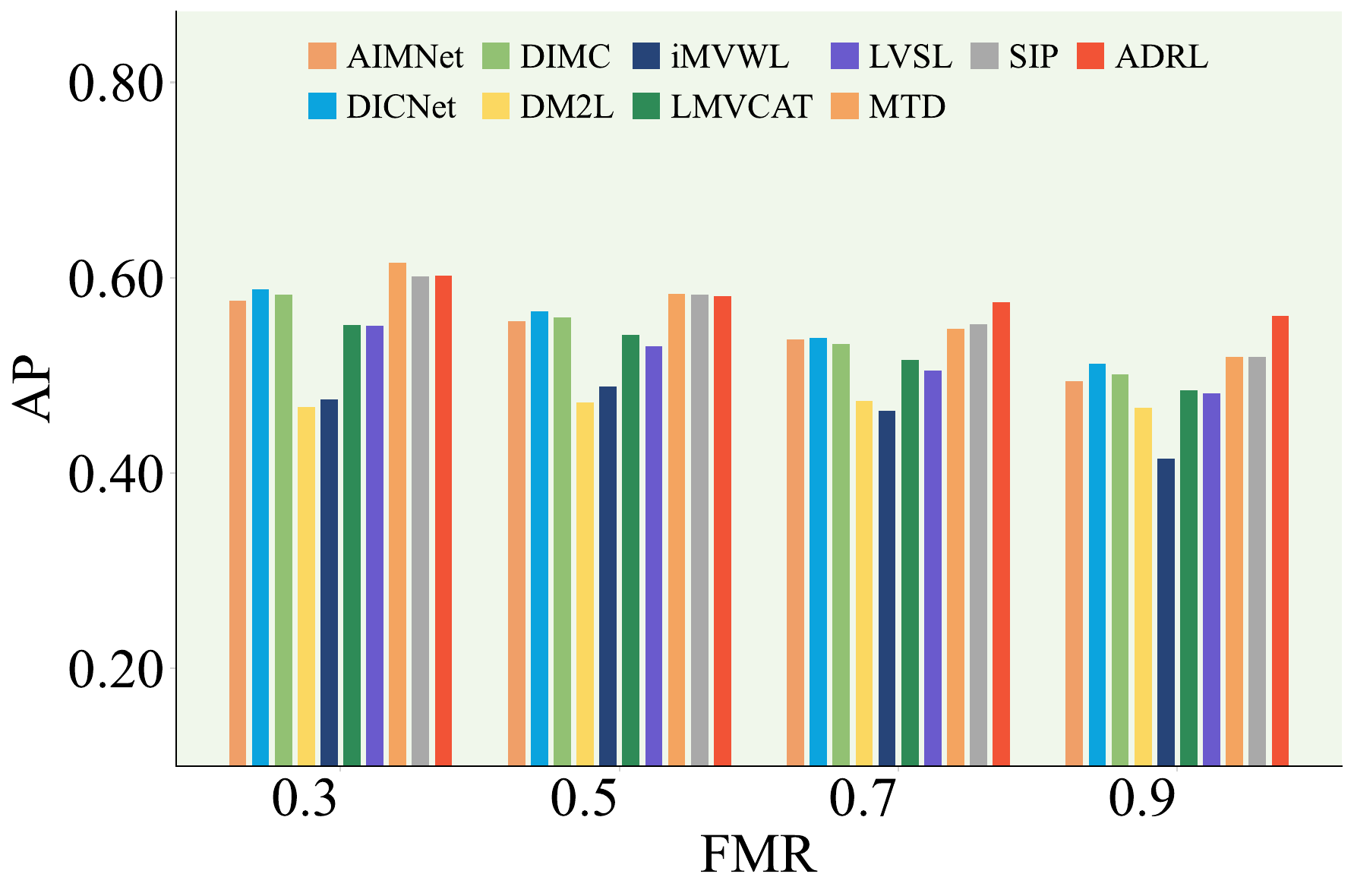}%
    }%
    \hfill % 水平填充
    \subfloat[Object\label{fig:image5}]{%
        \includegraphics[width=0.31\textwidth]{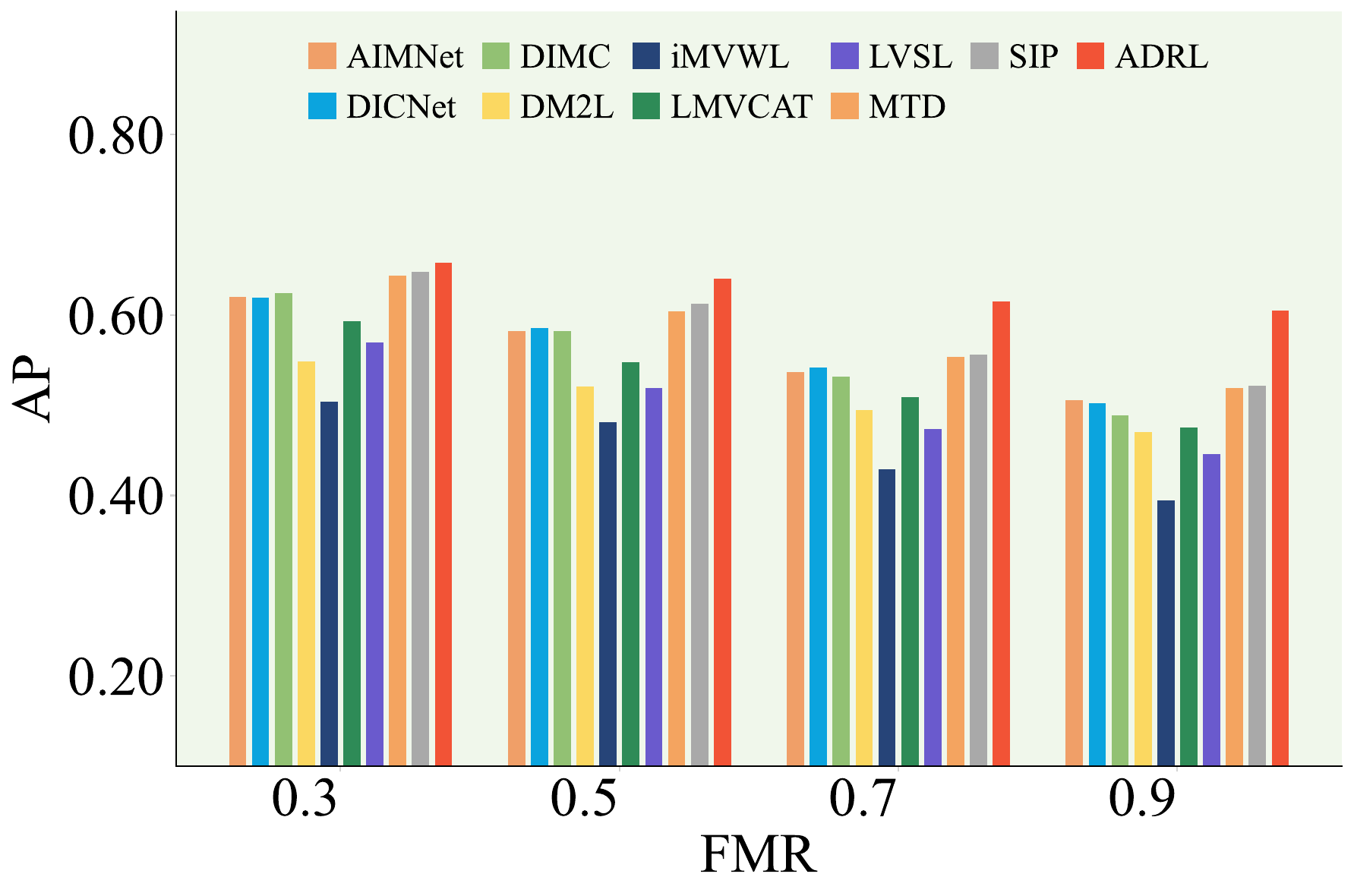}%
    }%
    \hfill % 水平填充
    \subfloat[Pascal07 \label{fig:image6}]{%
        \includegraphics[width=0.31\textwidth]{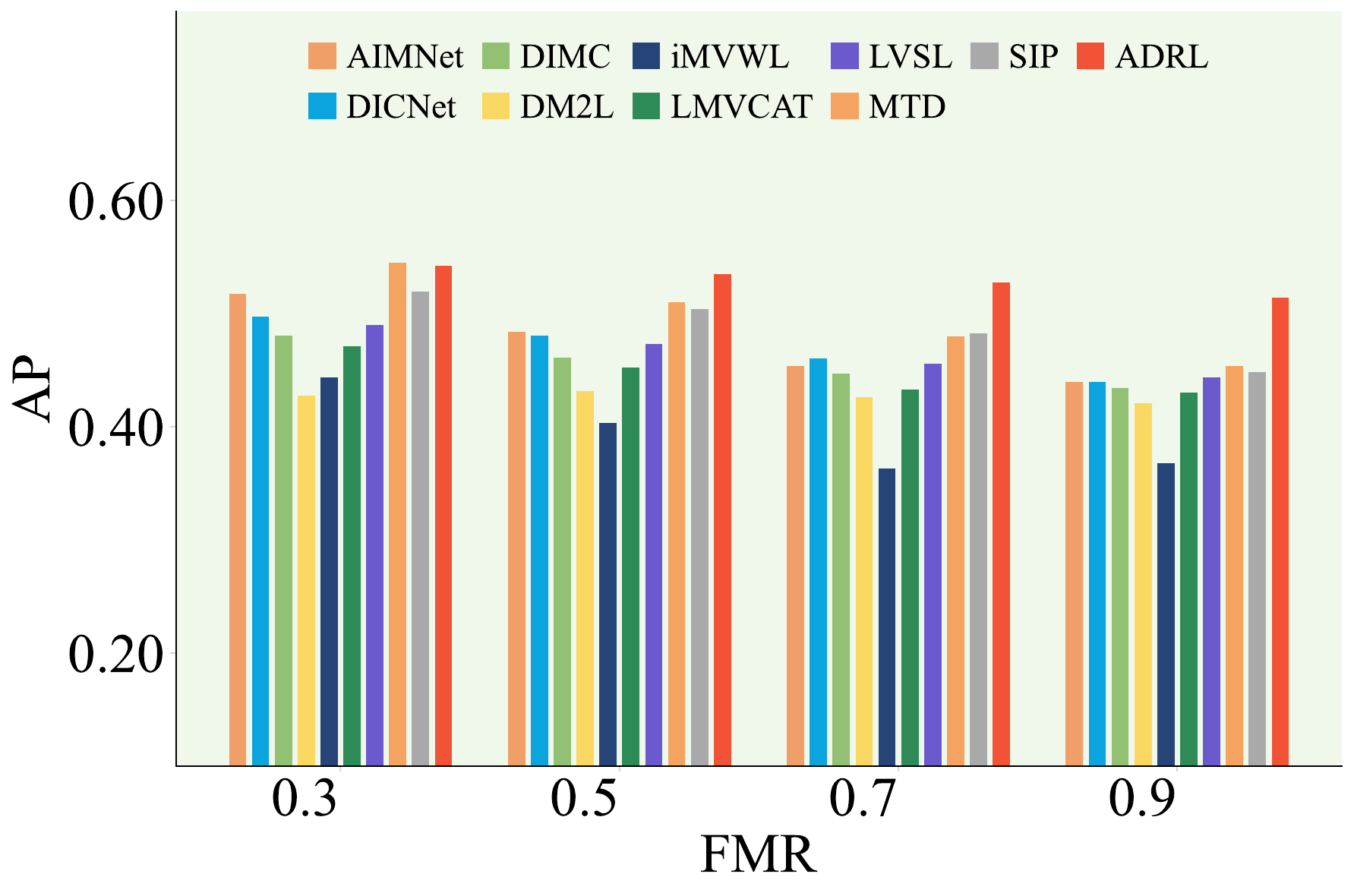}%
    }%

    \caption{Performance under different FMR with LMR fixed at 0.9.}
    \label{figIII}
\end{figure*}

\subsection{Experimental Setup}
\subsubsection{Datasets}

Following previous works \cite{liu2023dicnet,tan2018incomplete}, we evaluate our method on six widely-used multi-view multi-label datasets, i.e., 
Corel5k\cite{duygulu2002object}, Pascal07\cite{everingham2010pascal}, ESPGame\cite{von2004labeling}, IAPRTC12\cite{grubinger2006iapr}, Mirflickr\cite{huiskes2008mir}, 
and Object \cite{hao2024anchor}. In the first five datasets, six types of visual features are extracted to construct six separate views, i.e., GIST, HSV, Hue, SIFT, RGB, and LAB, while the last dataset comprises additional five views tailored for object recognition tasks. \textbf{Corel5k} includes 4,999 images, each annotated with 1 to 5 descriptive tags selected from a pool of 260 concepts.  \textbf{Pascal07} offers a benchmark collection of 9,963 images encompassing 20 object categories and is widely used for  detection tasks. \textbf{ESPGame} includes 20,770 images labeled via online games, with an average of 4.69 tags per image across 268 concepts. \textbf{IAPRTC12} is a large-scale dataset comprising 19,627 images across 291 categories, with each sample containing up to 23 labels extracted from slogans or subtitles. \textbf{ Mirflickr} contains 25,000 photographs sourced from the social photography site Flickr, with 38 curated tags used to support multi-label learning and evaluation. \textbf{Object} has 6047 instances requiring
 recognition, which are annotated with 31 attributes.

\subsubsection{Incomplete Dataset Preparation}
To evaluate the effectiveness of our approach under incomplete data scenarios, we replicate the  strategy introduced in \cite{liu2023dicnet,li2021concise} to transform fully observed datasets into incomplete multi-view multi-label versions. Specifically, to simulate partial views, a specified fraction of instances in each view is randomly masked according to the feature missing ratio (FMR). Importantly, we ensure that every instance retains at least one complete modality to maintain sample validity. To model weak supervision, a consistent proportion of both positive and negative labels across categories is randomly discarded, as determined by the label missing ratio (LMR).

\subsubsection{Comparison Methods}
To evaluate the performance of our ADRL, we conduct a comprehensive comparison with nine state-of-the-art methods, i.e., AIMNet\cite{liu2024attention}, DICNet\cite{liu2023dicnet}, DIMC\cite{wen2023deep}, DM2L\cite{ma2021expand}, iMVWL\cite{tan2018incomplete}, LMVCAT\cite{liu2023incomplete}, LVSL\cite{zhao2022non}, MTD\cite{liu2023masked}, and SIP\cite{liu2024partial}. It is important to note that LVSL  and DM2L are not designed to simultaneously handle incomplete views and missing labels. Consequently, to align them with our experimental conditions, we implement necessary adaptations.  Since LVSL is a MvMLC method that lacks the capability to process missing data, we address this limitation by imputing missing views using the average feature of the available instances. Besides, the unknown labels are filled with the value of ``0''. DM2L is specifically designed for learning from incomplete multi-label data. To enable the application of DM2L in a multi-view setting, we concatenate all recovered views into a unified representation. The parameters for each method are configured based on the recommended
values in their original source code.

\subsubsection{Evaluation Metrics}
In accordance with established evaluation protocols in multi-view multi-label learning \cite{liu2023masked}, the performance of our model is rigorously evaluated using six standard metrics, i.e., Ranking Loss (RL), Average Precision (AP), Hamming Loss (HL),  adapted area under curve (AUC),  OneError (OE) and Coverage (Cov). Besides, the values of 1-RL, 1-HL, 1-OE, and 1-Cov are reported to enable consistent comparison, such that higher scores uniformly indicate better performance.
To support holistic evaluation, the average ranking of each method across these six metrics is also computed.

\subsubsection{Implemention Details}
Each dataset is divided into training, validation and test sets in the ratio of 7:1:2. Stochastic Gradient Descent (SGD) optimizer is adopted for training with a  learning rate of 1.0 across all datasets.  The top-$k$ nearest neighbors for each sample are determined using the $k$-NN algorithm with the Euclidean distance metric, where k is set to 10. To enhance statistical reliability, we perform multiple repetitions of the random dataset splitting and the steps of incomplete data generation. The performance result is reported as the average value  together with the standard deviation. Our model is implemented by PyTorch on one NVIDIA GeForce RTX 4090 GPU of 24GB memory. 

\begin{table*}[htbp]
\centering
\adjustbox{max width=\textwidth,center}{
\begin{tabular}{ccccccccccc>{\columncolor{gray!15}}c}
\toprule
\textbf{DATA}   &
\textbf{METRIC} &
\textbf{AIMNet} &
\textbf{DICNet} &
\textbf{DIMC} &
\textbf{DM2L} &
\textbf{iMVWL} &
\textbf{LMVCAT} &
\textbf{LVSL} &
\textbf{MTD} &
\textbf{SIP} &
\textbf{ADRL} \\ \midrule
\multirow{6}{*}{\textbf{COR}} &
  \textbf{1-HL} &
  0.988(0.000) &
  0.987(0.000) &
  0.987(0.000) &
  0.987(0.000) &
  0.978(0.000) &
  0.986(0.000) &
  0.987(0.000) &
  0.988(0.000) &
  0.988(0.000) &
  \textbf{0.988(0.000)} \\
 &
  \textbf{1-OE} &
  0.478(0.011) &
  0.460(0.012) &
  0.446(0.009) &
  0.378(0.014) &
  0.308(0.017) &
  0.448(0.011) &
  0.353(0.017) &
  0.492(0.011) &
  0.492(0.014) &
  \textbf{0.521(0.007)} \\
 &
  \textbf{1-Cov} &
  0.766(0.004) &
  0.726(0.007) &
  0.709(0.008) &
  0.640(0.007) &
  0.701(0.003) &
  0.720(0.006) &
  0.720(0.005) &
  0.754(0.005) &
  0.781(0.005) &
  \textbf{0.797(0.007)} \\
 &
  \textbf{1-RL} &
  0.900(0.002) &
  0.881(0.004) &
  0.874(0.004) &
  0.843(0.004) &
  0.864(0.002) &
  0.876(0.004) &
  0.879(0.003) &
  0.893(0.004) &
  0.908(0.003) &
  \textbf{0.915(0.004)} \\
 &
  \textbf{AP} &
  0.404(0.005) &
  0.381(0.006) &
  0.370(0.005) &
  0.318(0.005) &
  0.281(0.005) &
  0.379(0.006) &
  0.311(0.005) &
  0.410(0.007) &
  0.414(0.006) &
  \textbf{0.438(0.006)} \\
 &
  \textbf{AUC} &
  0.903(0.002) &
  0.883(0.004) &
  0.877(0.004) &
  0.846(0.004) &
  0.867(0.002) &
  0.879(0.003) &
  0.882(0.002) &
  0.896(0.003) &
  0.910(0.002) &
 \textbf{ 0.917(0.003)} \\

 &
  \textbf{AVE} &
  3.50 &
  5.00 &
  7.17 &
  9.00 &
  9.50 &
  6.78 &
  7.33 &
  3.17 &
  2.17 &
  \textbf{1.00} \\ \midrule
\multirow{6}{*}{\textbf{ESP}} &
  \textbf{1-HL} &
  0.983(0.000) &
  0.983(0.000) &
  0.983(0.000) &
  0.983(0.000) &
  0.972(0.000) &
  0.982(0.000) &
  0.983(0.000) &
  0.983(0.000) &
  0.983(0.000) &
  \textbf{0.983(0.000)} \\
 &
  \textbf{1-OE} &
  0.442(0.006) &
  0.440(0.009) &
  0.431(0.009) &
  0.302(0.008) &
  0.343(0.010) &
  0.432(0.006) &
  0.365(0.006) &
  0.452(0.007) &
  0.450(0.006) &
  \textbf{0.478(0.011)} \\
 &
  \textbf{1-Cov} &
  0.621(0.003) &
  0.601(0.003) &
  0.586(0.004) &
  0.532(0.003) &
  0.548(0.004) &
  0.587(0.003) &
  0.578(0.002) &
  0.617(0.004) &
  0.622(0.004) &
  \textbf{0.646(0.007)} \\
 &
  \textbf{1-RL} &
  0.845(0.002) &
  0.836(0.002) &
  0.830(0.002) &
  0.804(0.002) &
  0.807(0.002) &
  0.827(0.002) &
  0.829(0.001) &
  0.843(0.002) &
  0.847(0.002) &
  \textbf{0.858(0.003)} \\
 &
  \textbf{AP} &
  0.306(0.003) &
  0.300(0.003) &
  0.294(0.003) &
  0.229(0.003) &
  0.243(0.004) &
  0.293(0.003) &
  0.266(0.003) &
  0.309(0.003) &
  0.309(0.004) &
  \textbf{0.329(0.006)} \\
 &
  \textbf{AUC} &
  0.850(0.001) &
  0.841(0.002) &
  0.835(0.002) &
  0.808(0.001) &
  0.813(0.002) &
  0.832(0.001) &
  0.834(0.001) &
  0.847(0.002) &
  0.851(0.002) &
  \textbf{0.862(0.003)} \\
 &
  \textbf{AVE} &
  3.67 &
  4.50 &
  5.67 &
  9.67 &
  9.13 &
  7.33 &
  7.17 &
  3.50 &
  2.33 &
  \textbf{1.00}\\ \midrule
  
\multirow{6}{*}{\textbf{IAP}} &
  \textbf{1-HL} &
  0.981(0.000) &
  0.981(0.000) &
  0.981(0.000) &
  0.980(0.000) &
  0.969(0.000) &
  0.980(0.000) &
  0.981(0.000) &
  0.981(0.000) &
  0.981(0.000) &
  \textbf{0.981(0.000)} \\
 &
  \textbf{1-OE} &
  0.457(0.008) &
  0.464(0.008) &
  0.454(0.006) &
  0.378(0.008) &
  0.351(0.008) &
  0.433(0.009) &
  0.377(0.007) &
  0.479(0.007) &
  0.459(0.005) &
  \textbf{0.483(0.012)} \\
 &
  \textbf{1-Cov} &
  0.675(0.004) &
  0.649(0.005) &
  0.630(0.005) &
  0.556(0.005) &
  0.565(0.004) &
  0.646(0.004) &
  0.605(0.004) &
  0.670(0.004) &
  0.678(0.003) &
  \textbf{0.703(0.006)} \\
 &
  \textbf{1-RL} &
  0.884(0.001) &
  0.874(0.002) &
  0.868(0.002) &
  0.837(0.002) &
  0.833(0.002) &
  0.868(0.002) &
  0.857(0.002) &
  0.882(0.002) &
  0.886(0.001) &
  \textbf{0.896(0.003)} \\
 &
  \textbf{AP} &
  0.329(0.003) &
  0.326(0.003) &
  0.318(0.002) &
  0.254(0.002) &
  0.236(0.002) &
  0.313(0.004) &
  0.262(0.002) &
  0.340(0.002) &
  0.331(0.003) &
  \textbf{0.349(0.008)} \\
 &
  \textbf{AUC} &
  0.885(0.001) &
  0.876(0.002) &
  0.870(0.001) &
  0.838(0.001) &
  0.835(0.001) &
  0.870(0.002) &
  0.859(0.001) &
  0.883(0.002) &
  0.887(0.001) &
  \textbf{0.897(0.002)} \\
 &
  \textbf{AVE} &
  4.00 &
  4.33 &
  6.00 &
  8.33 &
  9.78 &
  6.78 &
  8.00 &
  3.00 &
  2.78 &
  \textbf{1.00} \\ \midrule

\multirow{6}{*}{\textbf{MIR}} &
  \textbf{1-HL} &
  0.890(0.001) &
  0.890(0.001) &
  0.890(0.001) &
  0.876(0.001) &
  0.840(0.004) &
  0.880(0.004) &
  0.877(0.001) &
  0.893(0.001) &
  0.890(0.001) &
  \textbf{0.894(0.001)} \\
 &
  \textbf{1-OE} &
  0.646(0.009) &
  0.647(0.010) &
  0.646(0.008) &
  0.533(0.008) &
  0.512(0.016) &
  0.639(0.009) &
  0.609(0.007) &
  0.667(0.006) &
  0.654(0.007) &
  \textbf{0.681(0.008)} \\
 &
  \textbf{1-Cov} &
  0.673(0.003) &
  0.662(0.004) &
  0.657(0.003) &
  0.615(0.002) &
  0.588(0.013) &
  0.665(0.002) &
  0.624(0.002) &
  0.681(0.002) &
  0.669(0.006) &
  \textbf{0.692(0.002)} \\
 &
  \textbf{1-RL} &
  0.874(0.002) &
  0.869(0.003) &
  0.867(0.003) &
  0.835(0.001) &
  0.809(0.014) &
  0.862(0.003) &
  0.847(0.001) &
  0.878(0.001) &
  0.873(0.002) &
  \textbf{0.885(0.001)} \\
 &
  \textbf{AP} &
  0.599(0.003) &
  0.595(0.007) &
  0.592(0.006) &
  0.519(0.003) &
  0.495(0.017) &
  0.589(0.004) &
  0.548(0.003) &
  0.614(0.004) &
  0.603(0.005) &
  \textbf{0.628(0.006)} \\
 &
  \textbf{AUC} &
  0.861(0.001) &
  0.855(0.002) &
  0.854(0.002) &
  0.828(0.001) &
  0.801(0.017) &
  0.852(0.003) &
  0.839(0.001) &
  0.864(0.001) &
  0.859(0.002) &
  \textbf{0.872(0.002)} \\
 &
  \textbf{AVE} &
  3.78 &
  4.67 &
  6.17 &
  9.00 &
  10.00 &
  6.67 &
  8.00 &
  2.00 &
  3.50 &
  \textbf{1.00} \\
\midrule

\multirow{6}{*}{\textbf{OBJ}} &
  \textbf{1-HL} &
  0.948(0.001) &
  0.948(0.001) &
  0.947(0.001) &
  0.935(0.000) &
  0.899(0.002) &
  0.940(0.003) &
  0.935(0.001) &
  0.949(0.001) &
  0.948(0.001) &
  \textbf{0.951(0.002)} \\
 &
  \textbf{1-OE} &
  0.619(0.015) &
  0.601(0.011) &
  0.594(0.012) &
  0.537(0.011) &
  0.465(0.018) &
  0.604(0.016) &
  0.450(0.008) &
  0.627(0.011) &
  0.626(0.009) &
  \textbf{0.676(0.016)} \\
 &
  \textbf{1-Cov} &
  0.807(0.006) &
  0.794(0.006) &
  0.793(0.006) &
  0.768(0.005) &
  0.744(0.008) &
  0.796(0.008) &
  0.759(0.006) &
  0.813(0.006) &
  0.809(0.006) &
  \textbf{0.835(0.005)} \\
 &
  \textbf{1-RL} &
  0.888(0.005) &
  0.876(0.004) &
  0.875(0.004) &
  0.860(0.004) &
  0.833(0.006) &
  0.878(0.006) &
  0.850(0.005) &
  0.890(0.005) &
  0.889(0.004) &
  \textbf{0.910(0.004)} \\
 &
  \textbf{AP} &
  0.639(0.010) &
  0.627(0.009) &
  0.623(0.010) &
  0.577(0.009) &
  0.512(0.014) &
  0.630(0.012) &
  0.537(0.008) &
  0.649(0.009) &
  0.649(0.009) &
  \textbf{0.689(0.009)} \\
 &
  \textbf{AUC} &
  0.897(0.004) &
  0.886(0.004) &
  0.885(0.004) &
  0.872(0.004) &
  0.846(0.006) &
  0.888(0.006) &
  0.864(0.004) &
  0.900(0.005) &
  0.898(0.004) &
  \textbf{0.918(0.003)} \\
 &
  \textbf{AVE} &
  4.00 &
  5.67 &
  6.78 &
  8.17 &
  9.78 &
  5.33 &
  9.00 &
  2.00 &
  3.00 &
  \textbf{1.00} \\ \midrule
\multirow{6}{*}{\textbf{PAS}} &
  \textbf{1-HL} &
  0.932(0.001) &
  0.931(0.000) &
  0.931(0.001) &
  0.927(0.001) &
  0.882(0.004) &
  0.915(0.005) &
  0.928(0.001) &
  \textbf{0.933(0.001)} &
  0.932(0.001) &
  0.932(0.001) \\
 &
  \textbf{1-OE} &
  0.462(0.010) &
  0.443(0.007) &
  0.435(0.010) &
  0.419(0.006) &
  0.366(0.039) &
  0.433(0.017) &
  0.418(0.008) &
  0.474(0.008) &
  0.468(0.008) &
  \textbf{0.497(0.014)} \\
 &
  \textbf{1-Cov} &
  0.781(0.007) &
  0.749(0.003) &
  0.738(0.010) &
  0.720(0.004) &
  0.674(0.011) &
  0.759(0.006) &
  0.738(0.003) &
  0.790(0.006) &
  0.778(0.004) &
  \textbf{0.810(0.004)} \\
 &
  \textbf{1-RL} &
  0.830(0.006) &
  0.804(0.002) &
  0.792(0.008) &
  0.778(0.003) &
  0.736(0.011) &
  0.808(0.006) &
  0.797(0.002) &
  0.836(0.006) &
  0.828(0.004) &
 \textbf{0.856(0.004)} \\
 &
  \textbf{AP} &
  0.549(0.007) &
  0.517(0.004) &
  0.510(0.008) &
  0.482(0.005) &
  0.438(0.022) &
  0.524(0.009) &
  0.486(0.005) &
  0.562(0.005) &
  0.552(0.006) &
  \textbf{0.589(0.009)} \\
 &
  \textbf{AUC} &
  0.851(0.005) &
  0.827(0.002) &
  0.817(0.008) &
  0.806(0.003) &
  0.767(0.011) &
  0.830(0.006) &
  0.823(0.002) &
  0.855(0.006) &
  0.848(0.005) &
  \textbf{0.872(0.004)} \\
 &
  \textbf{AVE} &
  3.33 &
  5.67 &
  7.17 &
  8.67 &
  10.00 &
  6.00 &
  7.50 &
  1.78 &
  3.33 &
  \textbf{1.33} \\ \bottomrule

\end{tabular}
}
\caption{Experimental results of ten methods on the six datasets with 50\% missing instances and 50\% missing labels. ``Ave'' represents the average ranking of the corresponding method across all six metrics.}\label{labelIII}
\end{table*}

%%%%%%%%%%%%%%%%%%%%%%%%%%%%%%%%%%%%%%%%%%%%%%%%%%%%%%%%%%%%%%%%%%%%
\begin{figure*}[htbp] % figure* 表示图片跨双栏，htbp 是浮动参数
    \centering % 整体居中

    % --- 开始添加 Legend 图片 ---
    % 您需要将 "your_legend_image.pdf" 替换为您的 legend 图片的实际文件名和路径
    % width 可以根据您的 legend 图片的实际宽高比和期望的显示效果进行调整
    % 例如，0.8\textwidth, 0.9\textwidth, 或者 \textwidth (如果希望它撑满)
    \includegraphics[width=0.9\textwidth]{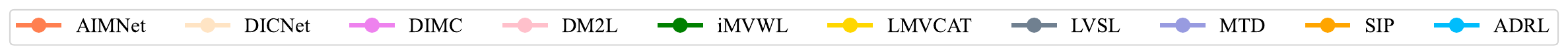} % <--- 在这里插入您的Legend图片

    % --- Legend 图片添加结束 ---

    % 第一行图片
    \subfloat[Corel5k\label{fig:image1}]{%
    \includegraphics[width=0.31\textwidth]{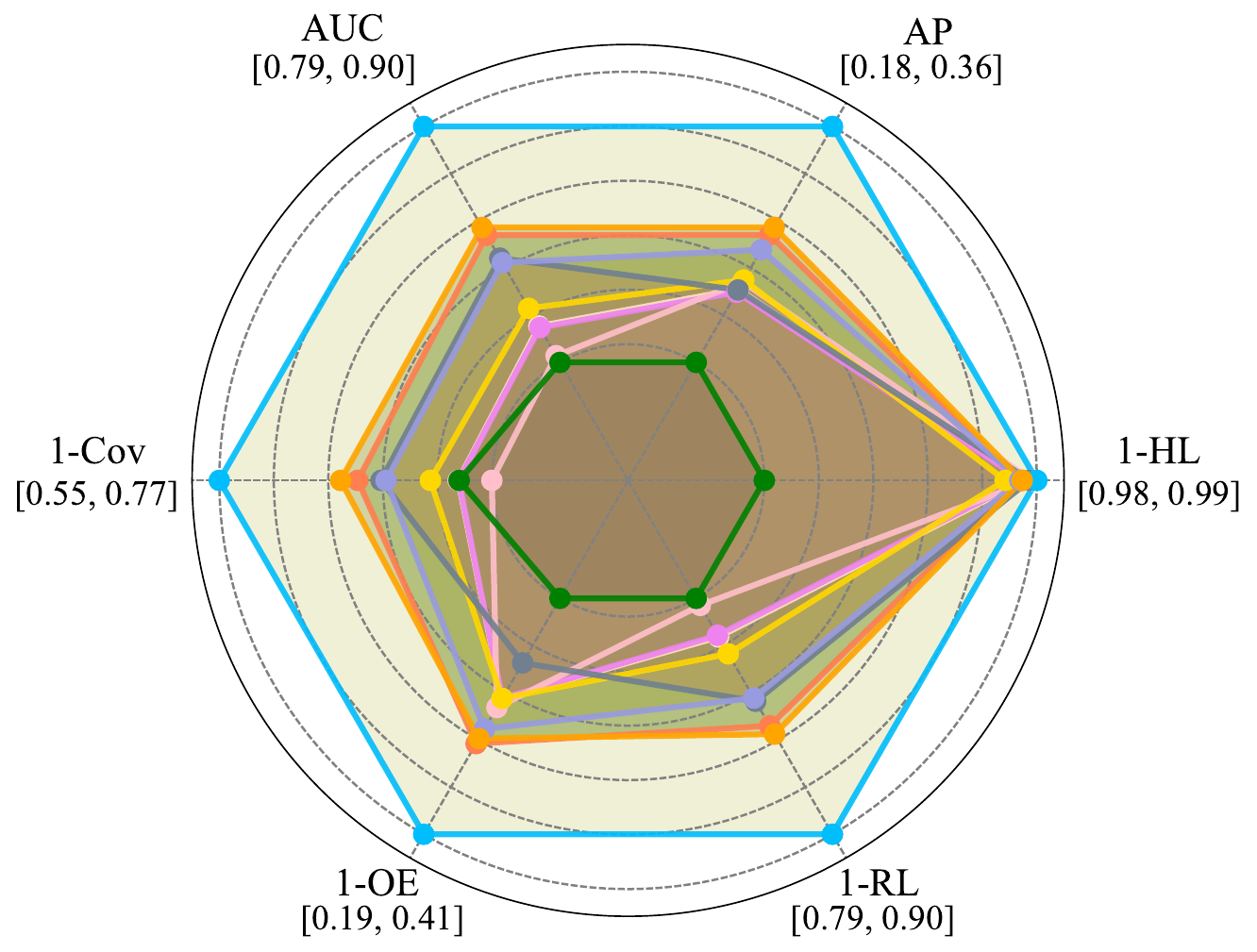}%
    }%
    \hfill % 水平填充，使图片之间有间隔
    \subfloat[ESPGame\label{fig:image2}]{%
        \includegraphics[width=0.31\textwidth]{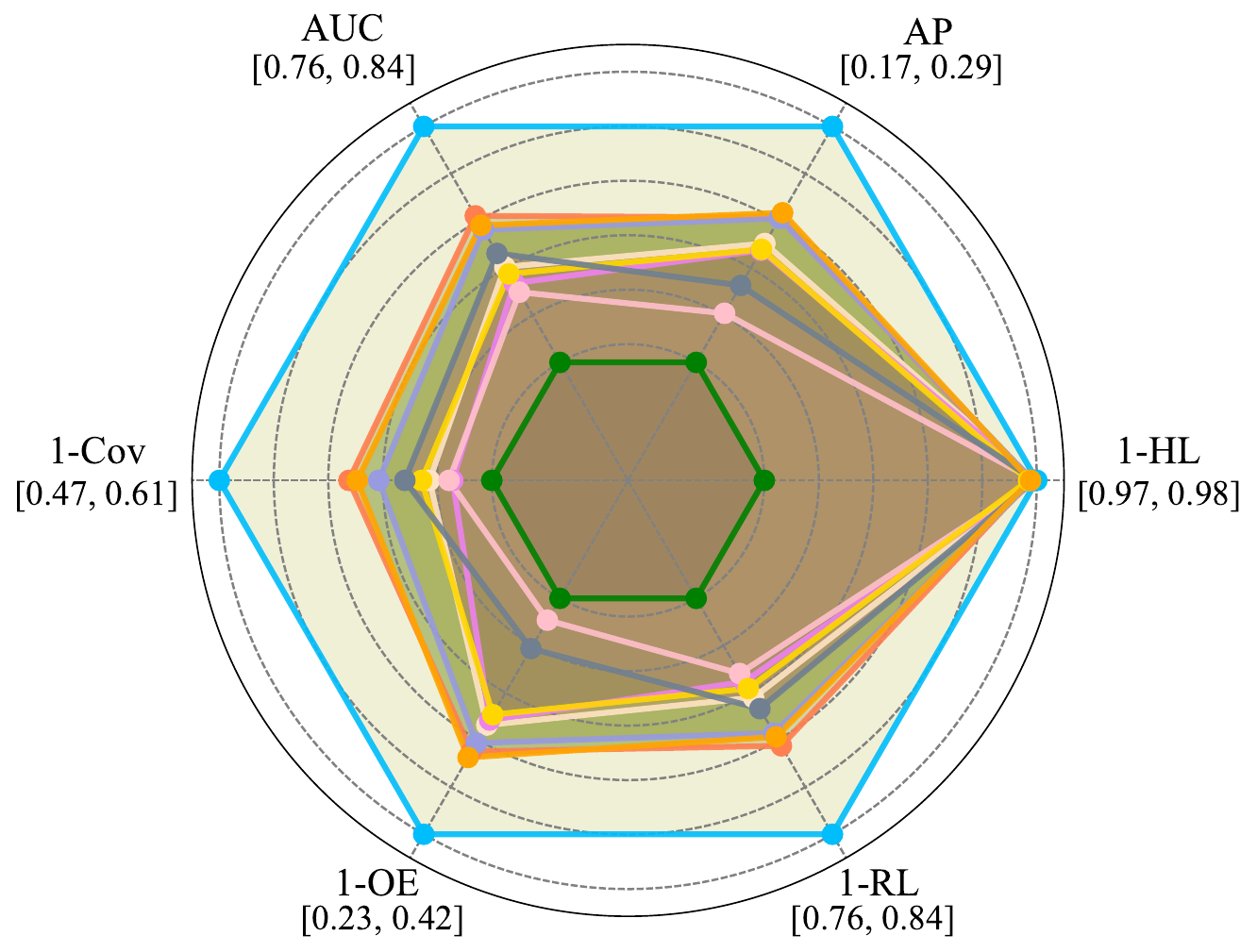}%
    }%
    \hfill % 水平填充
    \subfloat[IAPRTC12\label{fig:image3}]{%
        \includegraphics[width=0.31\textwidth]{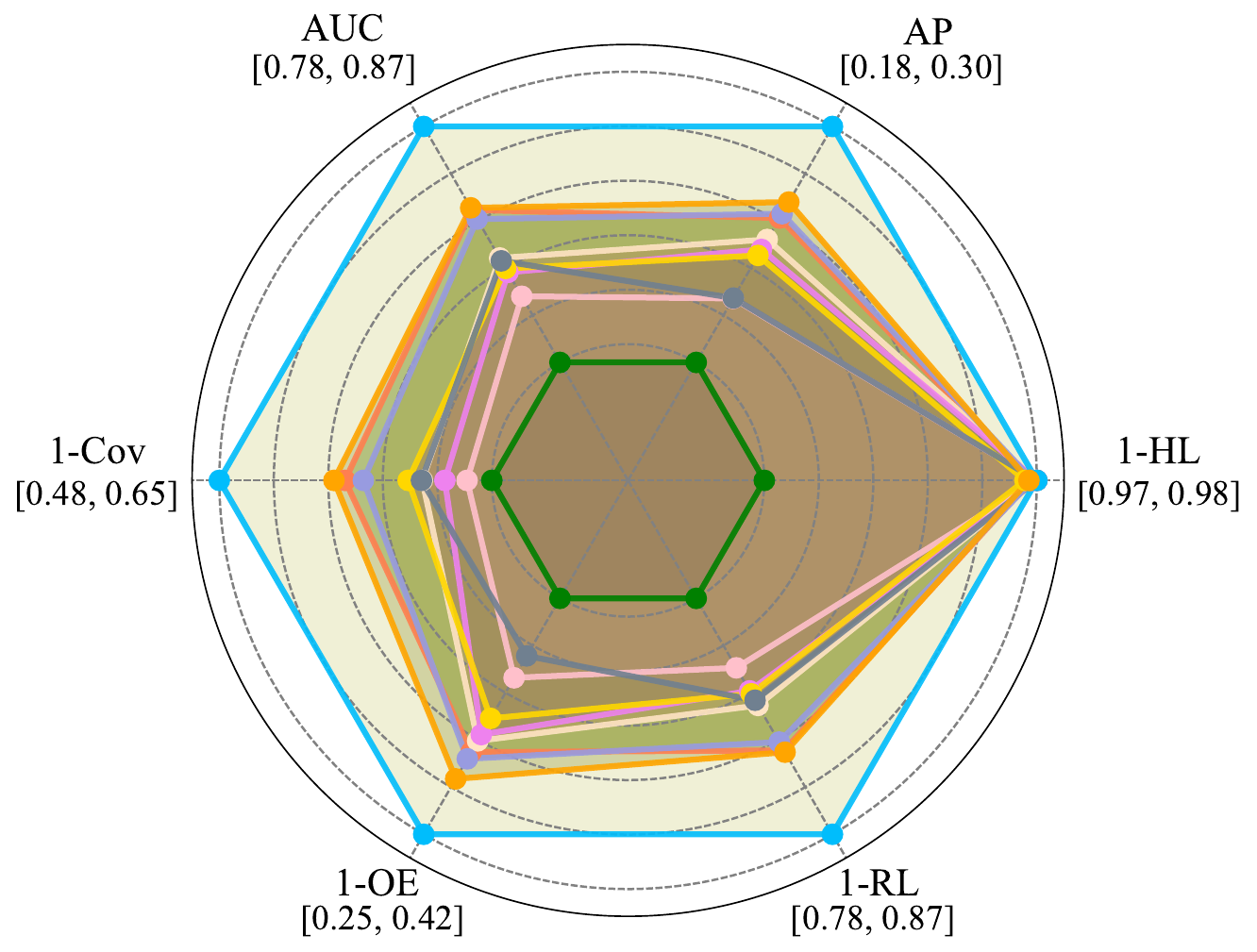}%
    }%
    \\ % 换行，开始新的一行子图

    % 第二行图片
    \subfloat[Mirflickr\label{fig:image4}]{%
        \includegraphics[width=0.31\textwidth]{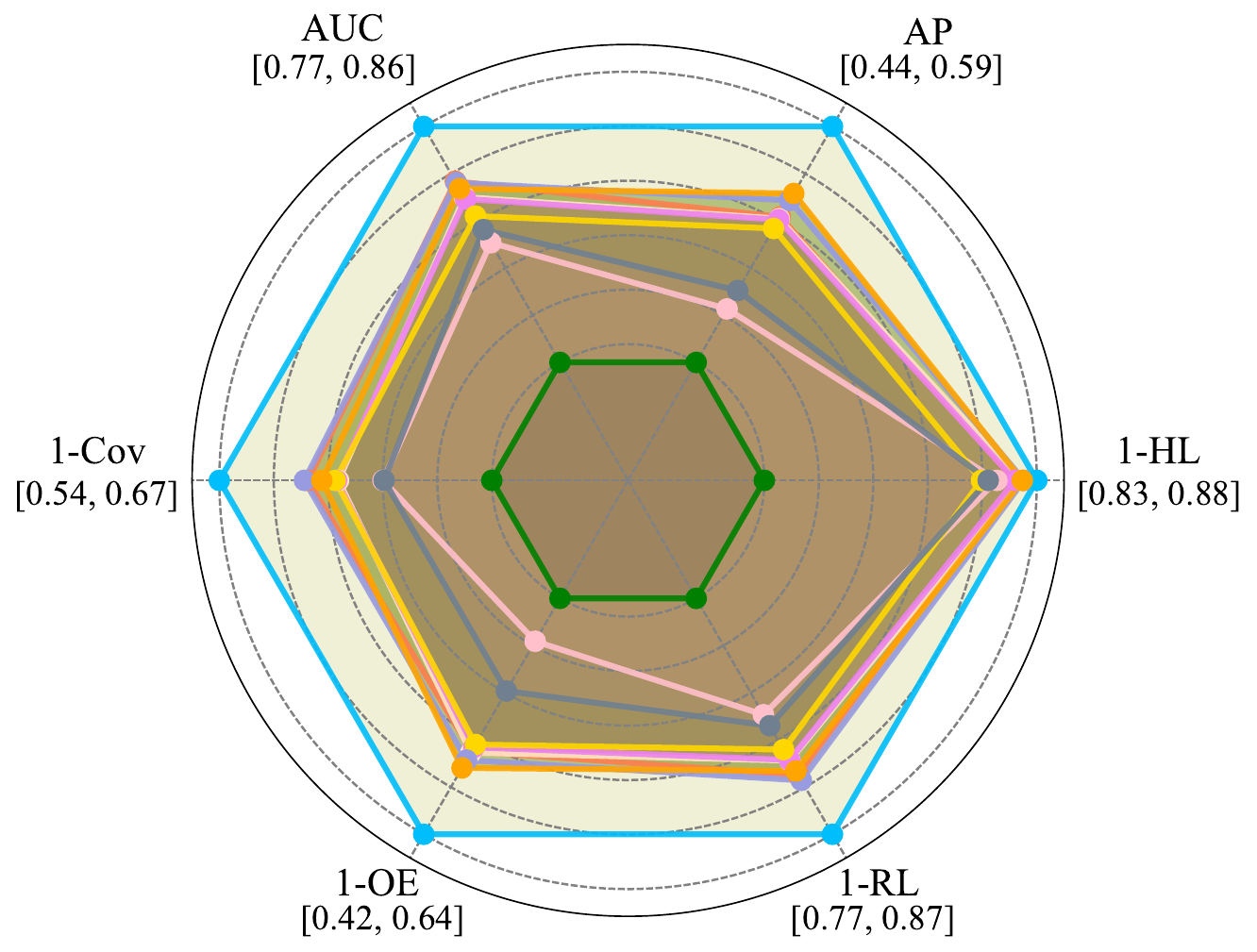}%
    }%
    \hfill % 水平填充
    \subfloat[Object\label{fig:image5}]{%
        \includegraphics[width=0.31\textwidth]{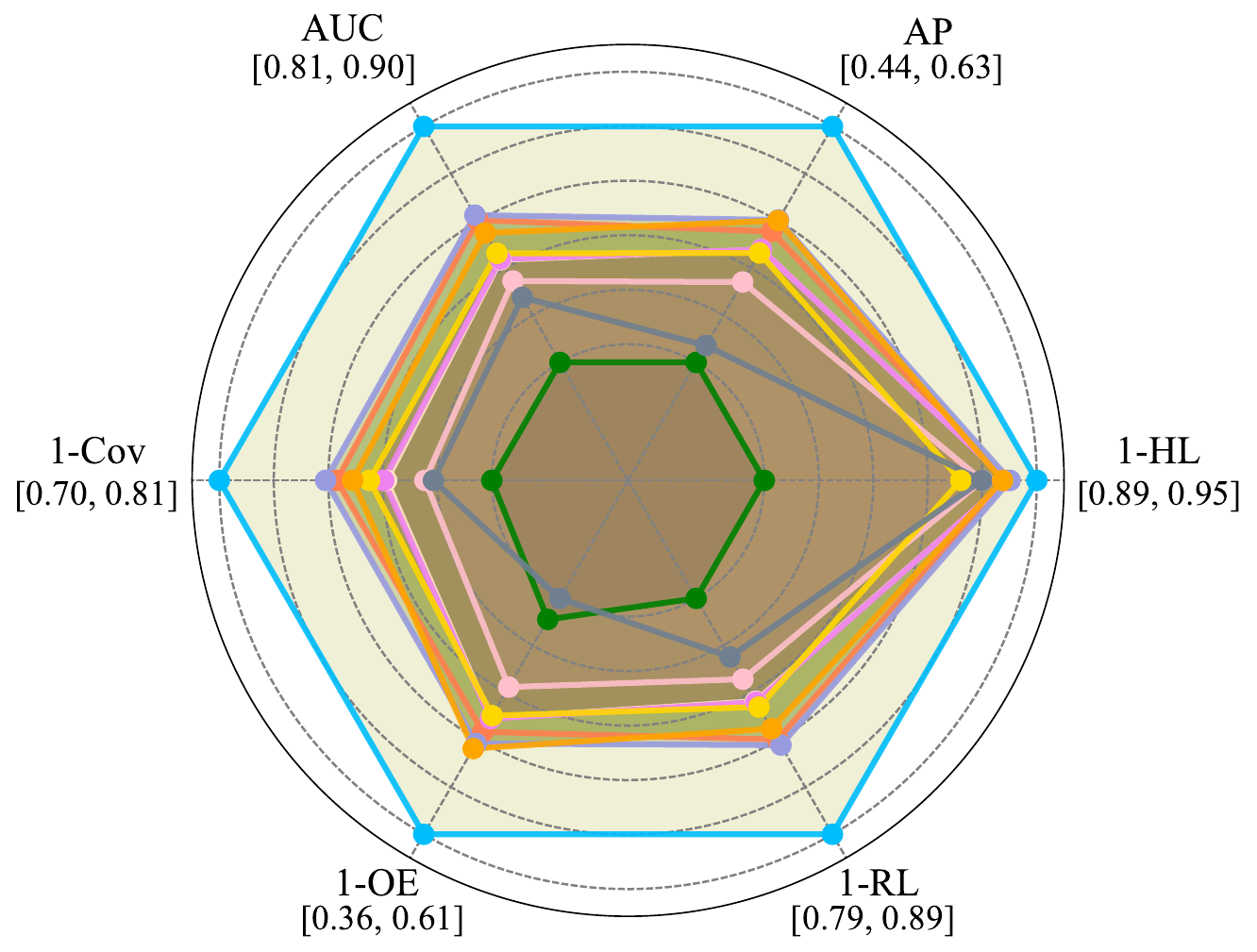}%
    }%
    \hfill % 水平填充
    \subfloat[Pascal07\label{fig:image6}]{%
        \includegraphics[width=0.31\textwidth]{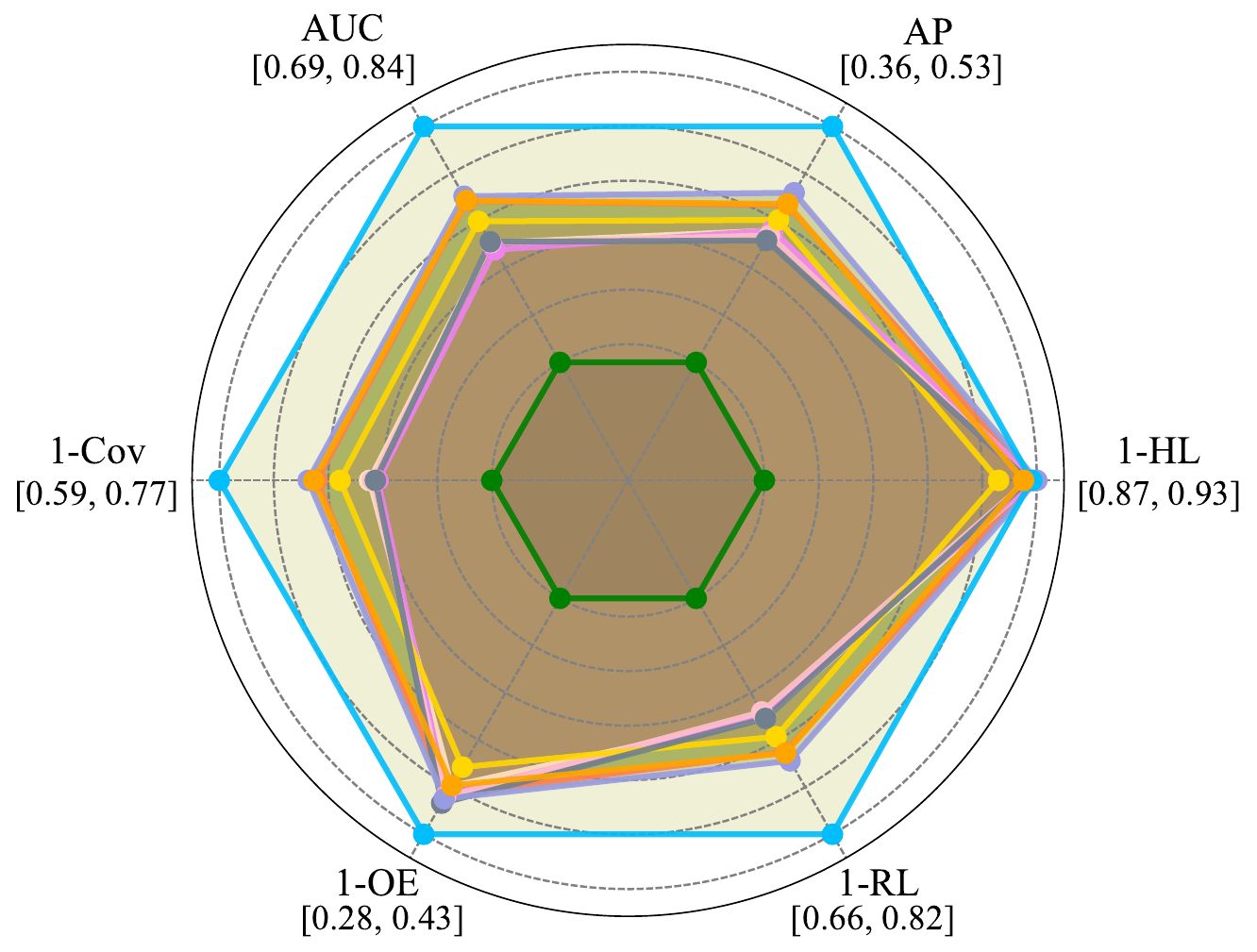}%
    }%
    \caption{Metric distributions under 70\% LMR and 90\% FMR.} % 您可能需要更新总标题以提及Legend
    \label{figIV} % 也可以更新总标签
\end{figure*}

\subsection{Experimental Results}
To examine how well our method addresses  the simultaneous problems of absent views and labels, we  perform extensive comparative experiments  against baseline algorithms across six datasets exhibiting varying degrees of data sparsity. The proportions of missing views (FMR) and labels (LMR) take values of $\{30\%,50\%,70\%,90\%\}$. The mean and standard deviation of the results with FMR and LMR fixed at 50\% is reported in Tabel \ref{labelIII}. Fig. \ref{figII} and \ref{figIII} illustrate the fluctuation in AP as PER and LMR vary between 30\% and 90\%. Moreover, in scenarios characterized by high levels of data missingness, i.e., LMR = 70\% and FMR = 90\%, the distributions of the six performance metrics are presented in Fig. \ref{figIV}. The results under alternative scenarios of pronounced data sparsity are provided in the supplementary material. Based on the comparison results, we have the following  observations:

1) As shown in Table \ref{labelIII}, our method demonstrates exceptional performance under missing data conditions, consistently emerging as the best-performing approach, as indicated by the boldfaced results in nearly every row. Except on the Pascal07 dataset, our method invariably ranks first, while the rankings of other methods remain unstable. Moreover, on the larger-scale datasets ESPGame and Mirflickr, ADRL outperforms the second-best approach SIP by a margin of 5\%.

2) From Fig. \ref{figII} and \ref{figIII}, it can be obtained that our method remains superior to all competitors under varying levels of missing data. ADRL is frequently one of the top performers when
 the missing ratio is low.  As the missing proportion increases,   the performance of ADRL declines within a controllable range on the datasets Corel5k and Pascal07, whereas the effectiveness of other methods is significantly compromised by data imperfections. Moreover, the widening performance gap underscores  that our method excels in  feature representation mining and label semantic learning, which ensures robust performance even with incomplete data.

3) As shown in Fig. \ref{figII}, our approach achieves an approximate 15\% performance gain on the datasets Corel5k and ESPGame  under 90\%  FMR and 90\% LMR. This substantial improvement highlights the resilience of our method under severe data incompleteness.  The  advantage is further demonstrated by the radar plots in Figs.
\ref{figIV}, where our method consistently occupies the outermost boundaries at high missing ratios. 
Notably, ADRL encloses  larger areas compared to other methods, affirming its effectiveness in high-deficiency settings and its potential for broader deployment.

4) Compared to conventional methods DM2L, iMVWL and LVSL, our approach achieves markedly superior performance, which underscores the pivotal role of obtaining   high-level representations
 in multi-view learning. Compared to the global weighting strategy employed by AIMNet, our proximity-sensitive reconstruction mechanism demonstrates greater stability. The advantage over SIP emphasizes the importance of extracting view-specific information, while the improvement over MTD highlights the necessity of adaptive modeling for complex inter-view interactions and the learning of label correlations.  

5) Based on the comprehensive evaluation system constructed from six metrics, our method shows competitive performance, with the results of  Hamming Loss being roughly equivalent to those of other approaches. Notably, it consistently ranks highest on discriminative metrics such as AP and AUC.  These results indicate that ADRL not only captures sparse positive labels with higher precision but also yields  effective label ranking across instances.

\subsection{Ablation Study}
\begin{table}[htbp]
    \centering
    \resizebox{\linewidth}{!}{
    \begin{tabular}{ccc|ccc|ccc}
    \toprule
    \multirow{2}{*}{$S_{1}$} & \multirow{2}{*}{$S_{2}$}& \multirow{2}{*}{$S_{3}$} & \multicolumn{3}{c|}{Corel5k}&
 \multicolumn{3}{c}{Object} \\
    \cmidrule{4-9}
     &  &  & AP & AUC & 1-RL & AUC  & AP & 1-RL\\
    \midrule   
    \XSolidBrush & \Checkmark & \Checkmark & 0.381  &0.894  & 0.891   &0.654 & 0.906 & 0.896 \\
    \Checkmark &  \XSolidBrush &  \Checkmark & 0.433   & 0.910 & 0.908 & 0.678 &0.912 & 0.904  \\
     \Checkmark & \Checkmark & \XSolidBrush &  0.423 & 0.915  &0.911  & 0.689&0.916&0.909 \\
      \Checkmark & \Checkmark & \Checkmark & \textbf{0.438}  & \textbf{0.920}   & \textbf{0.919}  &  \textbf{0.694} & \textbf{0.921} & \textbf{0.913} \\
    \bottomrule
    \end{tabular}
    }
    \caption{Ablation study on Yeast, Corel 5k and VOC 2007 with  LER=$6\%$. `\Checkmark' and `\XSolidBrush' represent the used and not used corresponding item, respectively.}\label{table:ablation}
\end{table}
 In this subsection, the ablation experiments are conducted to thoroughly examine the impact of the three core components of our ADRL, i.e., mask attention-guided missing-view completion $(S_1)$, disentangled representation learning for shared and
private features $(S_2)$, and label semantics learning and adaptive view fusion $(S_3)$. When $S_1$ is excluded, missing views are imputed by averaging the available ones.  Removing $S_2$ leads to the use of a single-channel multilayer perceptron for feature extraction, with losses $\mathcal{L}_{\text{re}}$ and $\mathcal{L}_{\text{Dis}}$ discarded. When $S_3$ is ablated, the model is unable to derive label prototypes and pseudo-labels. As an alternative, we perform weighted fusion of multi-view representations through  fully connected layers and apply a straightforward classifier for final prediction. From the ablation result presented in Table \ref{table:ablation}, we can know that all three components are indispensable to the model. Besides, the recovery mechanism plays a crucial role, as it enhances model functionality by providing additional feature information and ensures stability in subsequent representation extraction. Moreover, semantically distinct features demonstrate better performance compared to single-channel representations, while the integrated strategy of category-aware semantic learning and adaptive view fusion surpasses the effectiveness of basic view aggregation and independent classification of multiple labels. These findings underscore that our approach provides a comprehensive consideration of feature extraction, view fusion and label correlation learning.

 \subsection{Parameters Study}
\begin{figure}[htbp]
\centering\includegraphics[width=1\linewidth]{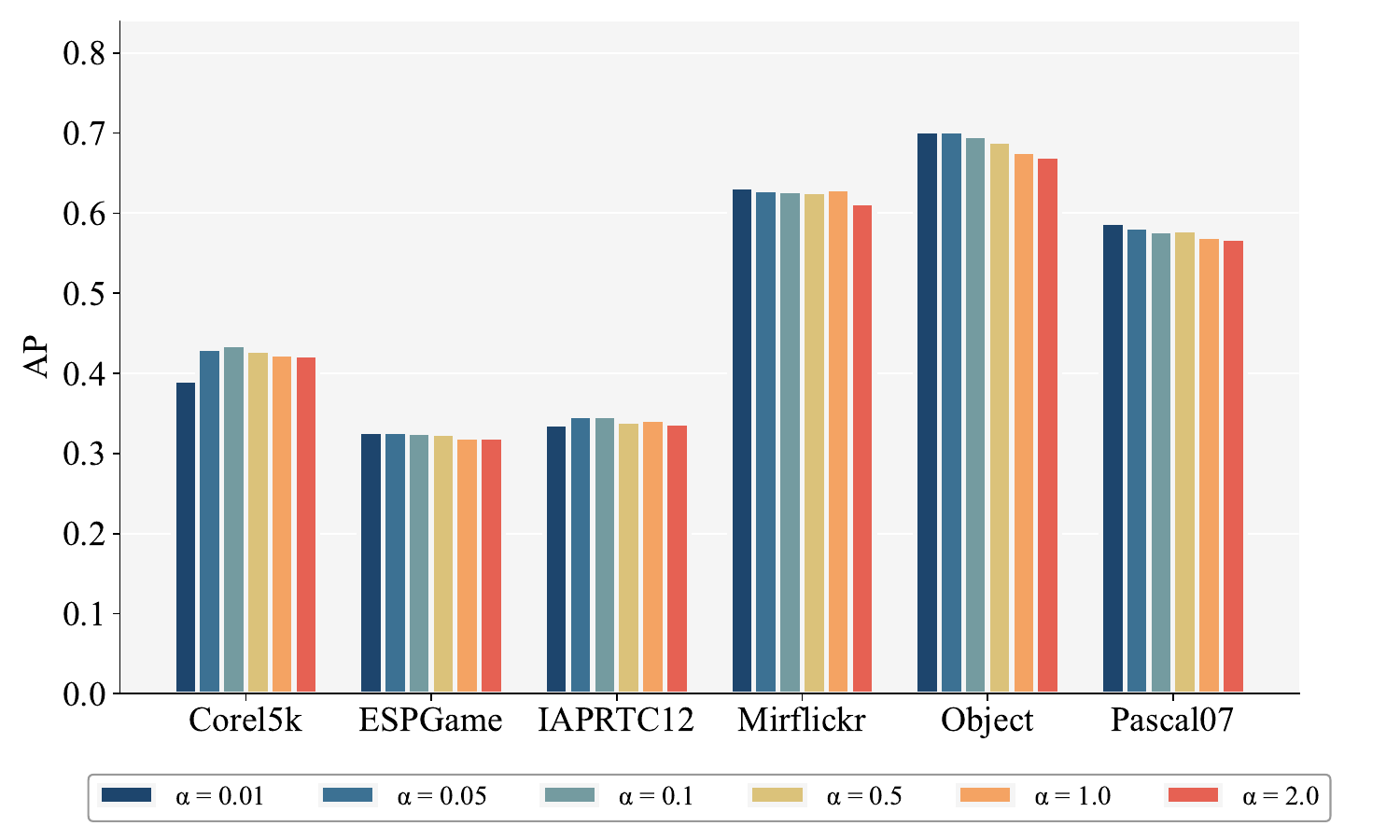}
    \caption{ Parameter study of $\alpha$.}
    \label{figa}
\end{figure}

\begin{figure}[htbp]
    \centering % 整个图表居中
    % 第一行图片
    \subfloat[Mirflickr]{\includegraphics[width=0.25\textwidth]{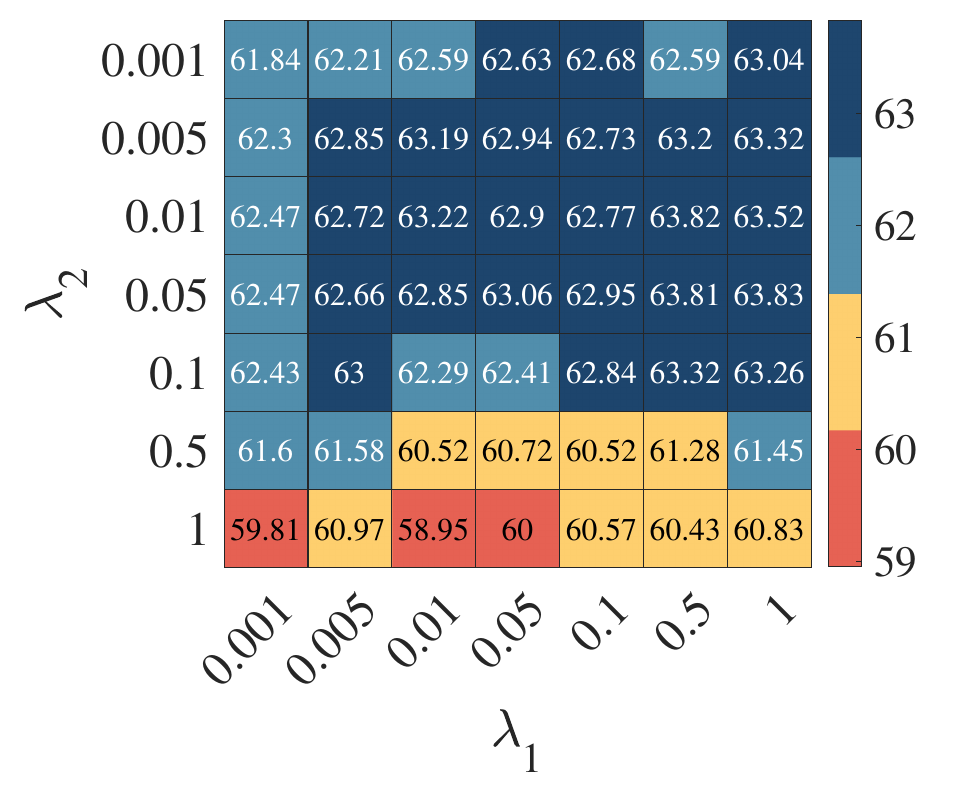}\label{fig:imgA}}
    % \hfill % 在第一张和第二张图之间插入弹性空间，将它们推开
    \subfloat[Object]{\includegraphics[width=0.25\textwidth]{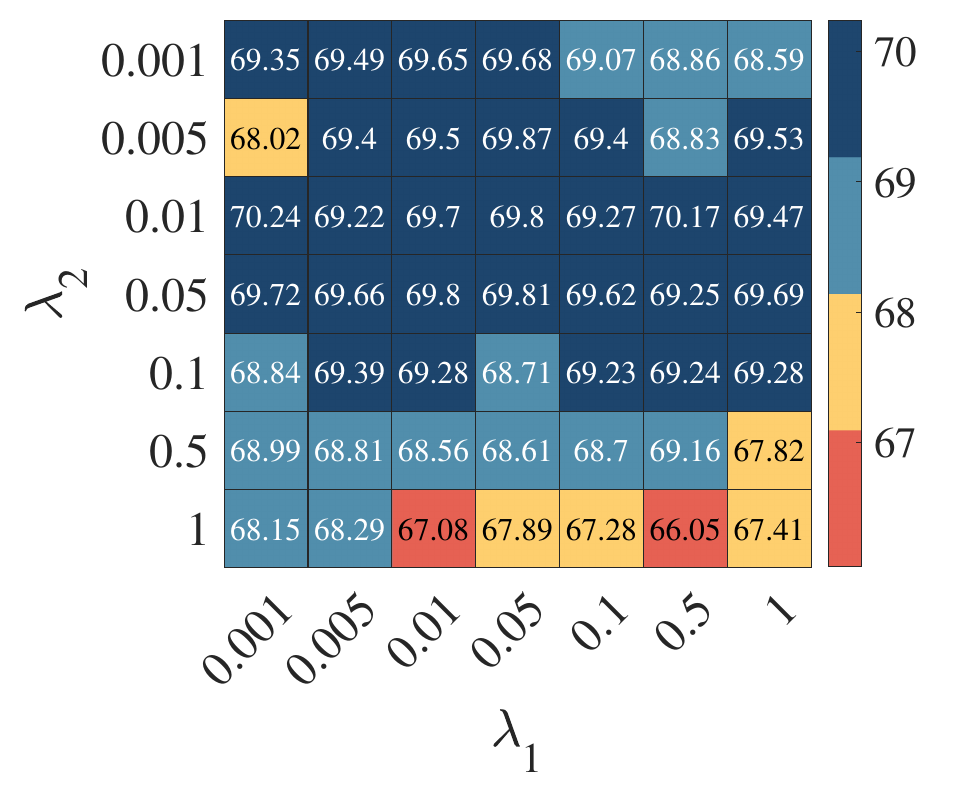}\label{fig:imgB}}
    % % 换行并减小垂直间距，让两行图片更紧凑
    % % -1ex 可以根据需要调整，负值越大，间距越小
    \\
    % 第二行图片
    \subfloat[Corel5k]{\includegraphics[width=0.25\textwidth]{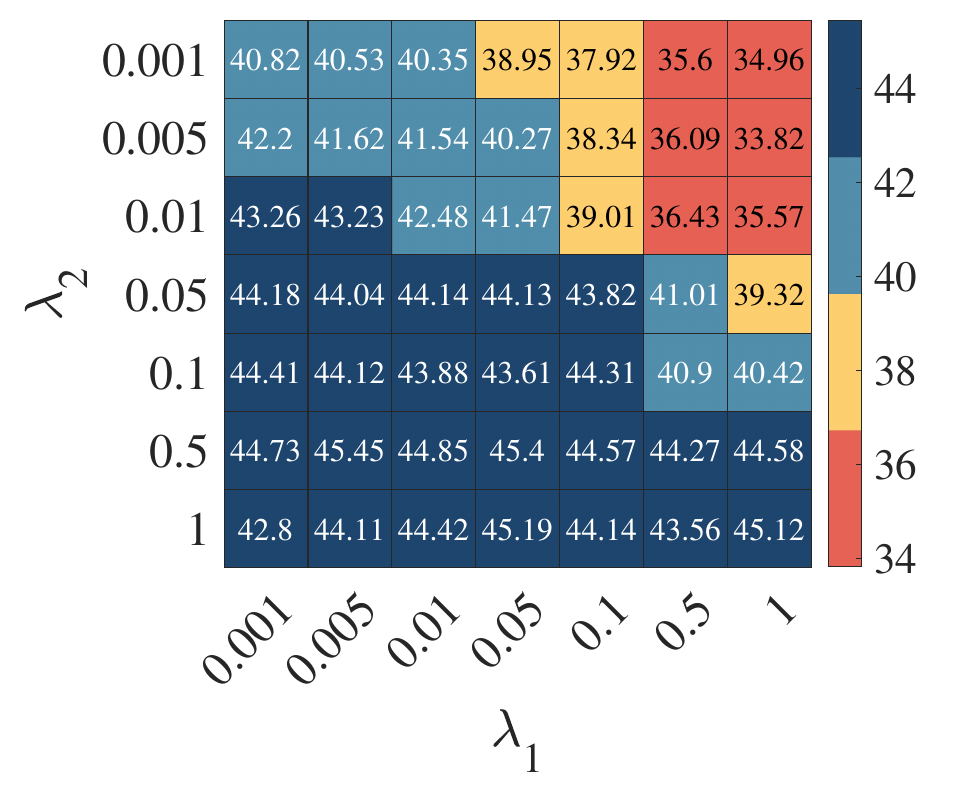}\label{fig:imgC}}
    % \hfill % 在第三张和第四张图之间插入弹性空间
    \subfloat[Pascal07]{\includegraphics[width=0.25\textwidth]{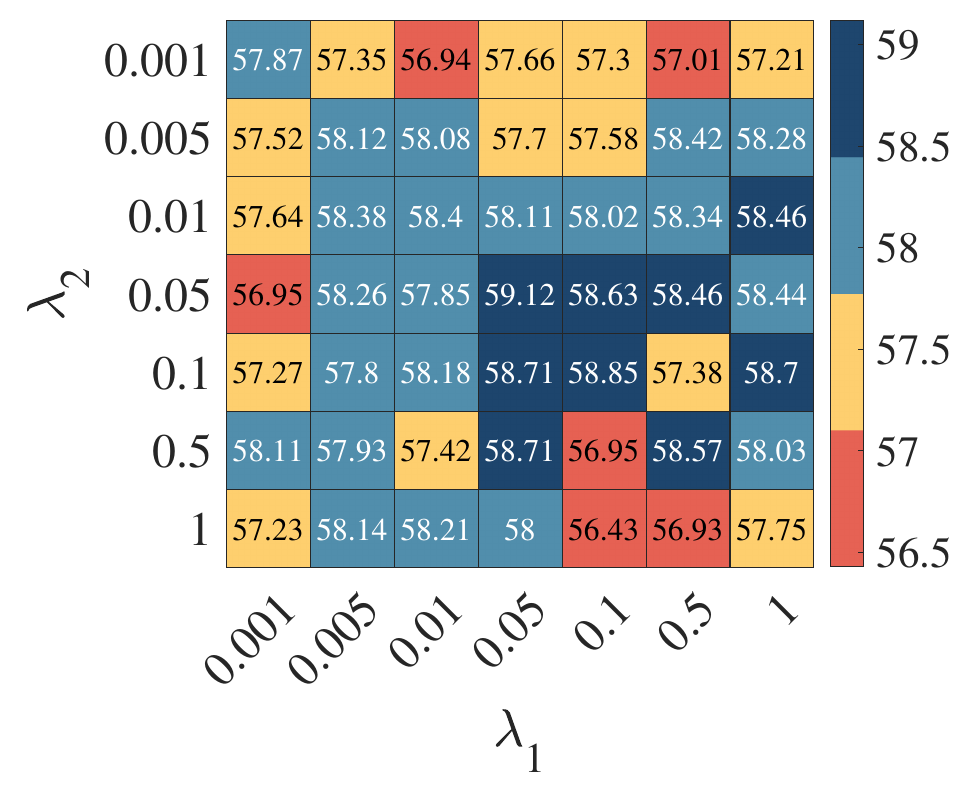}\label{fig:imgD}}
    % 主标题
    \caption{Parameter study of $\lambda_1$ and $\lambda_2$.}
    \label{figtwo}
\end{figure}

In our ADRL framework, there are three hyper-parameters, i.e.,  $\alpha$, $\lambda_1$ and $\lambda_2$ that require fine-tuning.  To evaluate the influence of these hyper-parameters on the classification  performance, we alter their settings  under 50\% FMR  and 50\% LMR, and observe the corresponding changes in AP. Given that $\alpha$ is responsible for controlling the weight of the overall classification loss, an insufficient value may lead to significant performance degradation. Then, our experimental results indicate that the model achieves optimal performance when $\alpha$ is selected from a relatively constrained range of $(0.01, 2)$, as illustrated in Fig. \ref{figa}. Besides, the model exhibits low sensitivity to $\alpha$ within this interval. The remaining two hyper-parameters $\lambda_1$ and $\lambda_2$ are employed to balance the contributions of the reconstruction loss and the classification loss on pseudo-labels, respectively.  The joint impact of  $\lambda_1$ and $\lambda_2$ is explored by adjusting them over \{0.001, 0.005, 0.01, 0.05, 0.1, 0.5, 1\}, with the outcomes summarized in Fig. \ref{figtwo}. From the heatmap, we can find that  the optimal performance is achieved when $\lambda_1$ falls within the interval $(0.05, 1)$.
As for  $\lambda_2$, the most favorable results are observed in the range $(0.001, 0.05)$ for the datasets Mirflickr and Object, whereas for  Corel5k and Pascal07, peak performance is associated with the interval $(0.05, 1)$.

\subsection{Convergence Analysis}

To thoroughly assess the convergence characteristics of our proposed ADRL, we illustrate the simultaneous evolution of the training loss and the performance metric on the validation set during the training process. As depicted in Fig. \ref{figVV}, the value of the objective function decreases monotonically with successive iterations,  ultimately converging to a stable state. Concurrently, AP score demonstrates a continuous upward trajectory as the number of epochs grows. This phenomenon indicates that our model exhibits desirable convergence property, benefiting from a well-designed loss function that effectively guides the learning process toward an optimal solution. Moreover, the number of classes appears to be a key factor affecting the convergence rate. For datasets containing relatively few classes,
such as  Mirflickr and Pascal07,  the training process converges swiftly and  typically reaches stability within fewer than ten epochs. In contrast, other datasets tend to undergo a more gradual convergence process.

\begin{figure}[htbp] % 使用 figure (没有星号)，图片会浮动在单栏内
    \centering % 建议将内容居中在栏内

    \subfloat[Corel5k\label{fig:singlecol_img1}]{%
        % 将 "your_image1.pdf" 替换为您的图片文件名和路径
        % width=0.48\columnwidth 表示图片宽度为当前栏宽的48%
        % 两个0.48\columnwidth 加上 \hfill 应该能很好地并排放在一栏内
        \includegraphics[width=0.48\columnwidth]{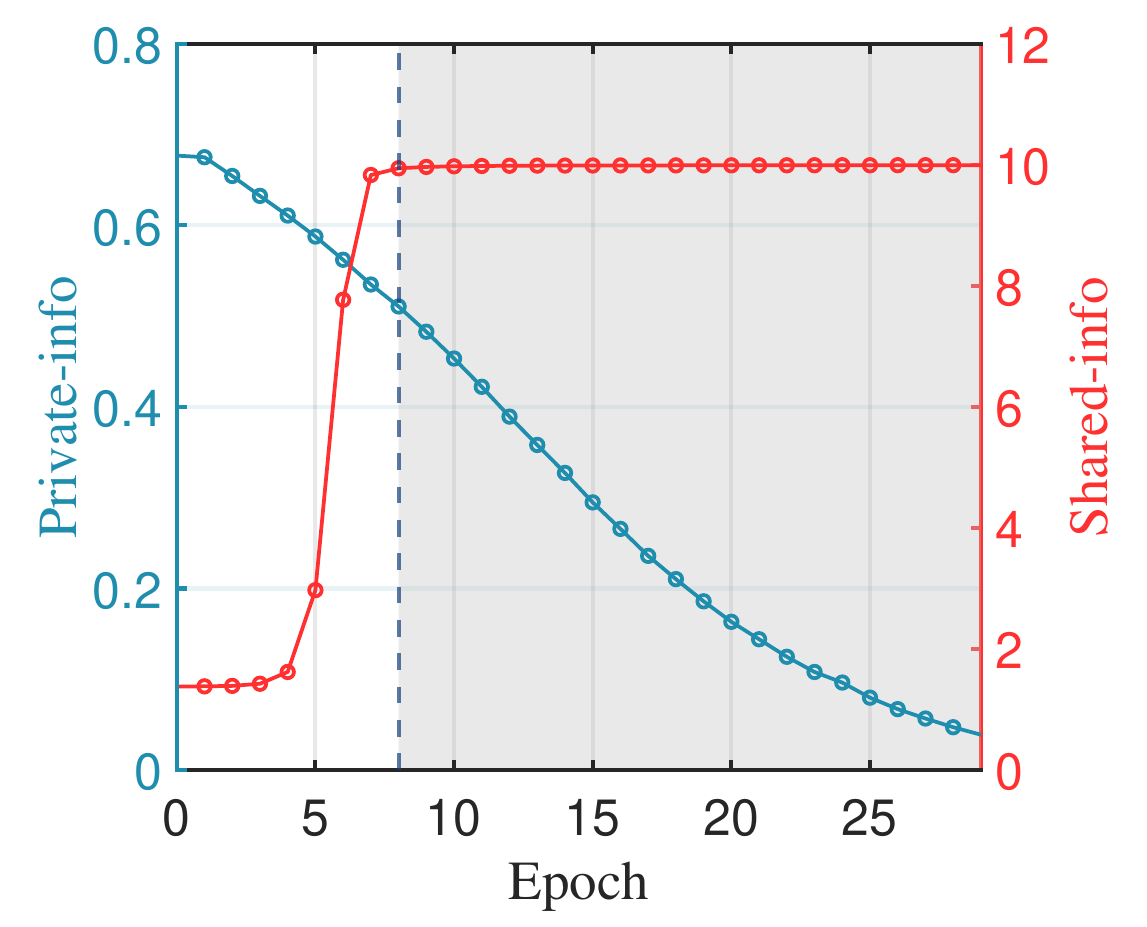}%
    }
    \hfill % 水平填充，在两张图片之间创建间隔
    \subfloat[Pascal07\label{fig:singlecol_img2}]{%
        % 将 "your_image2.pdf" 替换为您的图片文件名和路径
        \includegraphics[width=0.48\columnwidth]{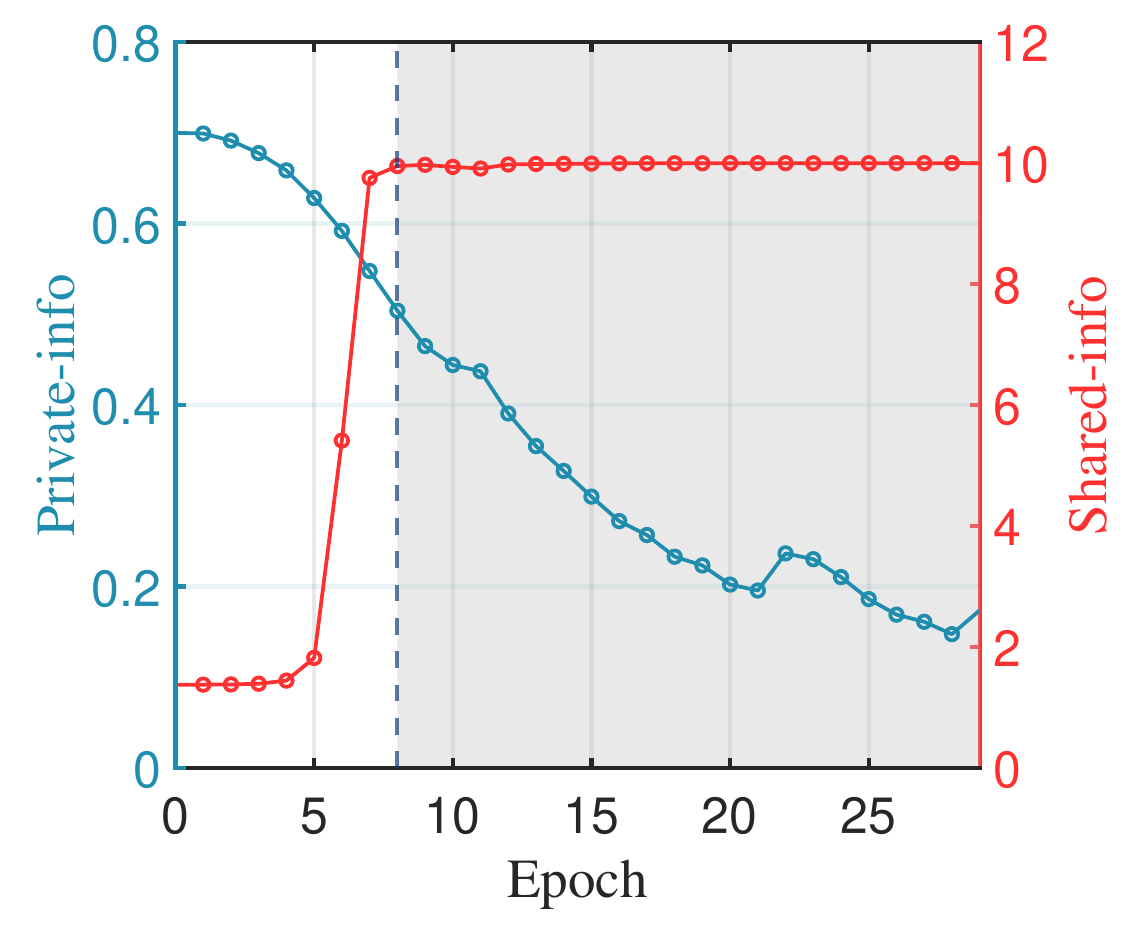}%
    }
    \caption{ Changes in mutual information among shared and specific representations.}
    \label{fig09}
\end{figure}

\begin{figure}[htbp] % figure* 表示图片跨双栏，htbp 是浮动参数
    \centering % 整体居中

    % 第一行图片
    \subfloat[Corel5k\label{converimage1}]{%
        \includegraphics[width=0.24\textwidth]{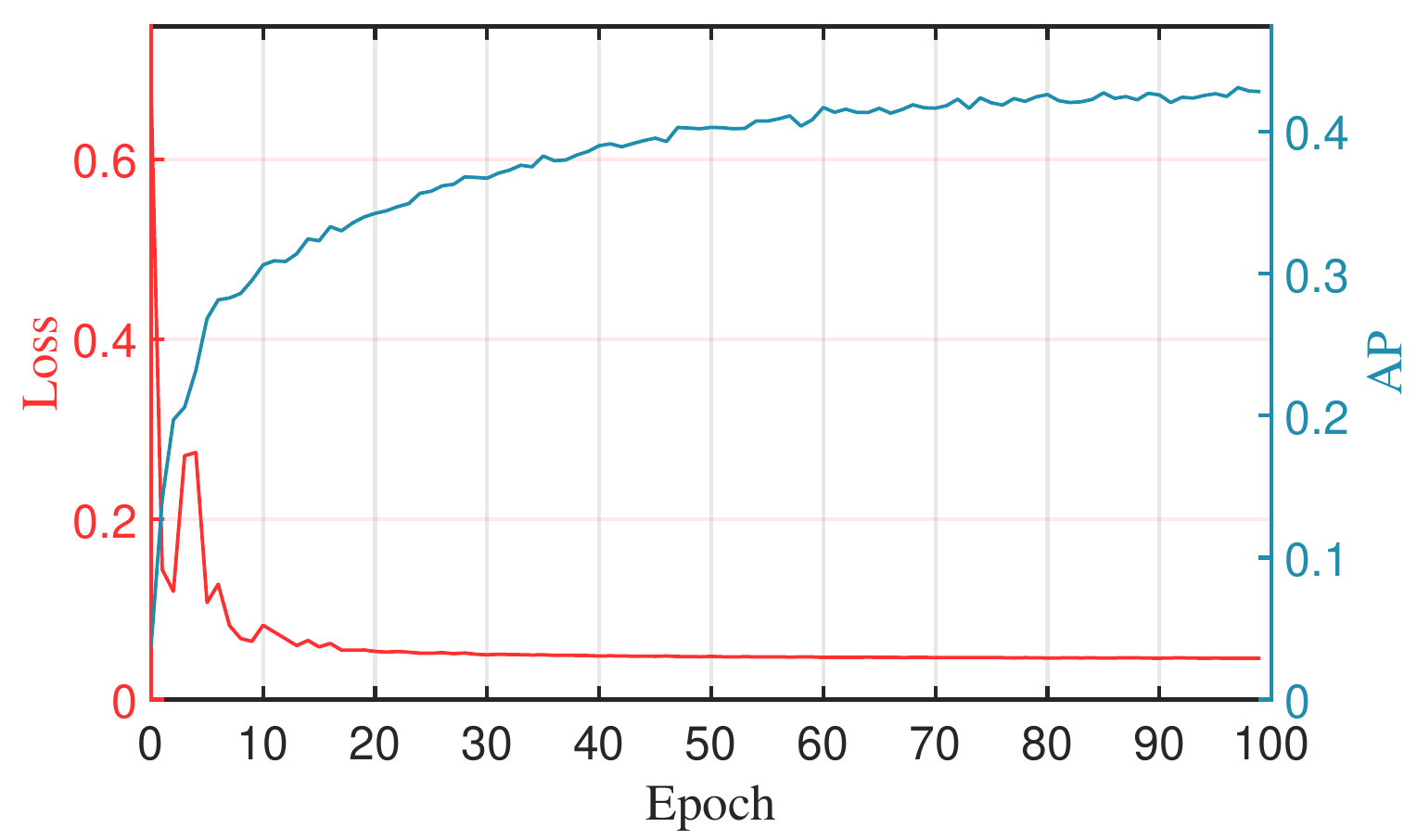}%
    }%
    \hfill % 水平填充，使图片之间有间隔
    \subfloat[ESPGame\label{converimage2}]{%
        \includegraphics[width=0.24\textwidth]{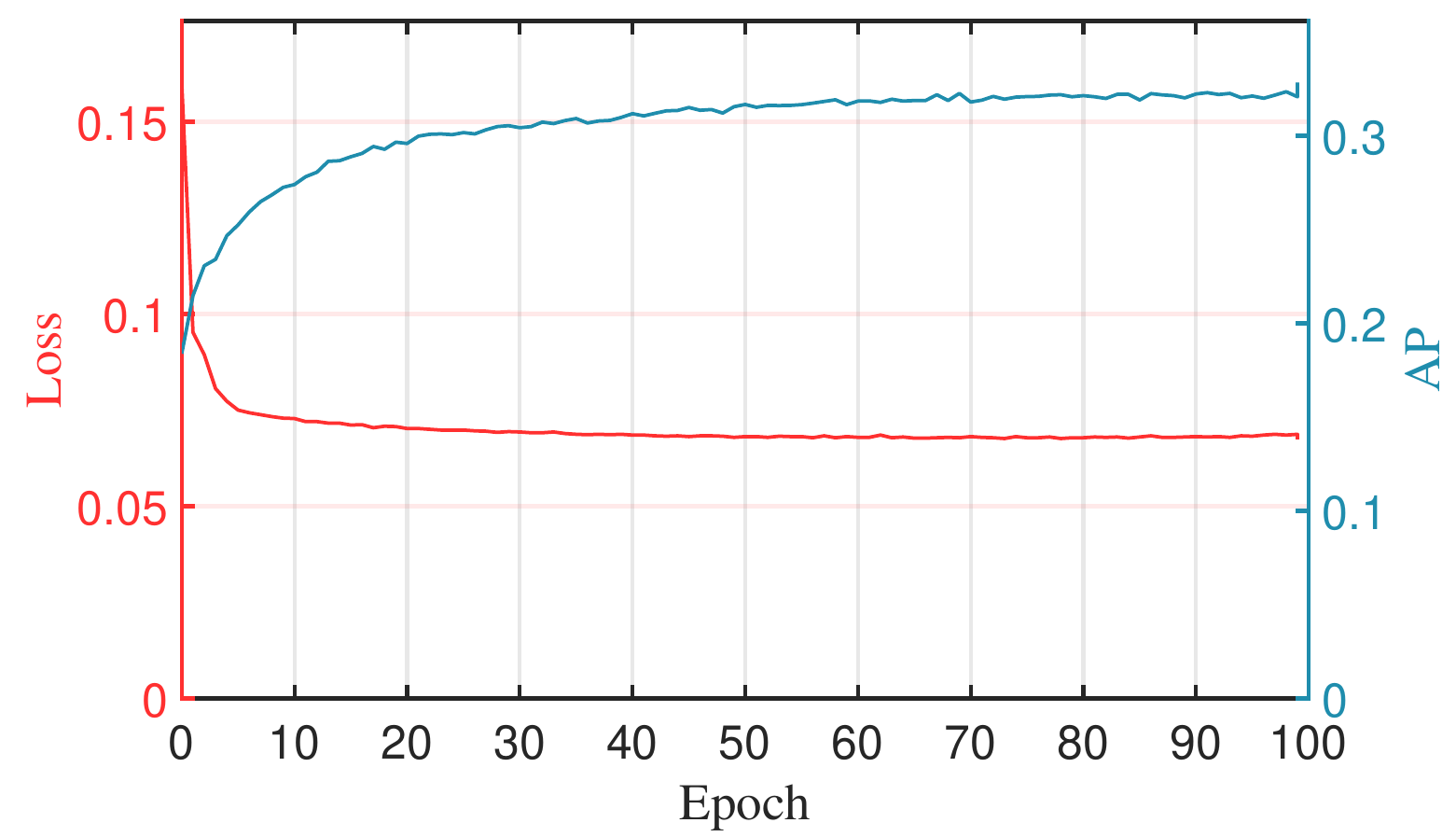}%
    }%
    \\ % 换行，开始新的一行子图
    % 第二行图片
    \subfloat[Mirflickr\label{converimage4}]{%
        \includegraphics[width=0.24\textwidth]{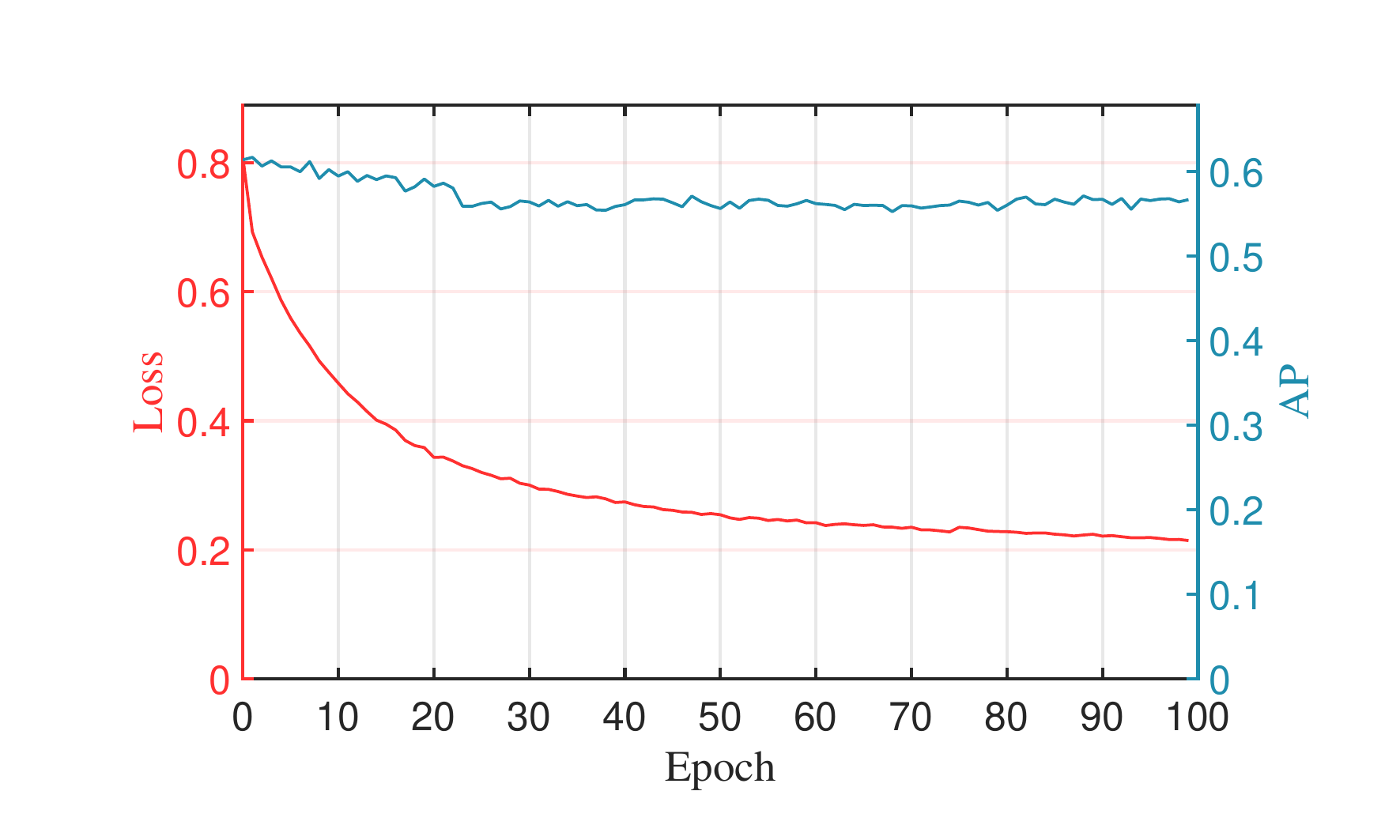}
    }%
    \hfill % 水平填充
    \subfloat[Pascal07\label{converimage6}]{%
        \includegraphics[width=0.24\textwidth]{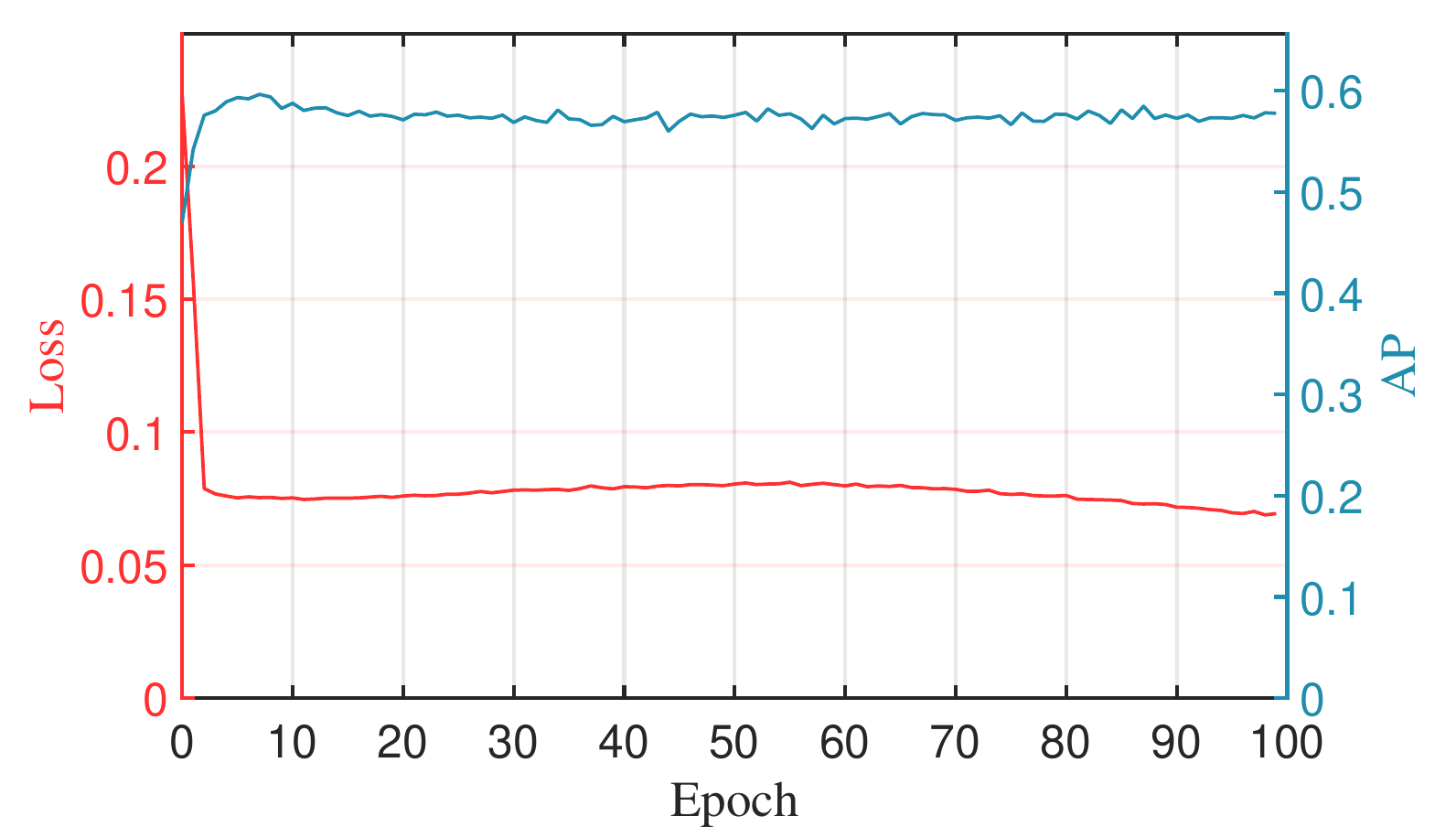}%
    }%
    \caption{Convergence Study.} % 您可能需要更新总标题以提及Legend
    \label{figVV} % 也可以更新总标签
\end{figure}

Another  important aspect of convergence worth examining is whether the semantic information of the shared and specific representations is disentangled during the training phase.   To explore this behavior, we conduct convergence analysis on the  datasets Corel5k and Pascal07 by tracking the evolution of mutual information among representations. From Fig. \ref{fig09}, two key observations can be made: 1) The shared information increases rapidly and stabilizes within 50 epochs, which indicates effective alignment and information exchange among shared representations. 2) Mutual information between specific representations drops steadily and remains consistently low, suggesting that information redundancy is well suppressed and  the unique characteristics  of each view are prominently captured. Thus, the trend shown in  Fig. \ref{fig09} provides  evidence that our model effectively separates representation semantics. 

% 在您希望插入图片的地方(例如在某个段落后)

\subsection{Application to Potential Prediction of NBA Players}
\renewcommand{\arraystretch}{1.5} % 调整行距
\begin{table*}[t!]
    \centering
    \label{tab:nba_comparison}
    \small 
    \begin{adjustbox}{width=\textwidth, center}
    \begin{tabular}{l|cccc|cccc|cccc}
        \hline \hline
        \multirow{2}{*}{\textbf{Method}} 
        & \multicolumn{4}{c|}{FMR=50\% and LMR=50\%} 
        & \multicolumn{4}{c|}{FMR=70\% and LMR=70\%} 
        & \multicolumn{4}{c}{FMR=90\% and LMR=90\%} \\
        \cline{2-13}
        & \textbf{1-HL} & \textbf{1-RL} & \textbf{AP} & \textbf{AUC} 
        & \textbf{1-HL} & \textbf{1-RL} & \textbf{AP} & \textbf{AUC} 
        & \textbf{1-HL} & \textbf{1-RL} & \textbf{AP} & \textbf{AUC} \\
        \hline
        \textbf{AIMNet} & $0.898_{0.001}$ & $0.910_{0.001}$ & $0.683_{0.002}$ & $0.922_{0.001}$ & $0.891_{0.002}$ & $0.890_{0.001}$ & $0.635_{0.004}$ & $0.904_{0.001}$ & $0.875_{0.001}$ & $0.860_{0.001}$ & $0.572_{0.001}$ & $0.877_{0.001}$ \\
        \textbf{DICNet} & $0.898_{0.001}$ & $0.906_{0.001}$ & $0.673_{0.003}$ & $0.917_{0.002}$ & $0.892_{0.001}$ & $0.888_{0.002}$ & $0.631_{0.005}$ & $0.901_{0.003}$ & $0.883_{0.000}$ & $0.867_{0.002}$ & $0.580_{0.003}$ & $0.881_{0.002}$ \\
        \textbf{DIMC} & $0.899_{0.001}$ & $0.909_{0.001}$ & $0.681_{0.005}$ & $0.920_{0.001}$ & $0.892_{0.002}$ & $0.889_{0.001}$ & $0.633_{0.004}$ & $0.902_{0.001}$ & $0.882_{0.002}$ & $0.864_{0.001}$ & $0.574_{0.005}$ & $0.880_{0.001}$ \\
        \textbf{DM2L} & $0.883_{0.000}$ & $0.861_{0.000}$ & $0.585_{0.002}$ & $0.877_{0.001}$ & $0.882_{0.000}$ & $0.848_{0.001}$ & $0.554_{0.003}$ & $0.865_{0.001}$ & $0.882_{0.000}$ & $0.808_{0.004}$ & $0.505_{0.003}$ & $0.827_{0.004}$ \\
        \textbf{iMVWL} & $0.830_{0.001}$ & $0.844_{0.003}$ & $0.527_{0.013}$ & $0.860_{0.003}$ & $0.829_{0.001}$ & $0.842_{0.001}$ & $0.522_{0.004}$ & $0.858_{0.001}$ & $0.823_{0.001}$ & $0.834_{0.001}$ & $0.510_{0.004}$ & $0.850_{0.001}$ \\
        \textbf{LMVCAT} & $0.890_{0.003}$ & $0.908_{0.001}$ & $0.683_{0.002}$ & $0.921_{0.001}$ & $0.886_{0.002}$ & $0.888_{0.003}$ & $0.633_{0.005}$ & $0.903_{0.003}$ & $0.872_{0.001}$ & $0.859_{0.001}$ & $0.565_{0.001}$ & $0.876_{0.001}$ \\
        \textbf{LVSL} & $0.882_{0.000}$ & $0.871_{0.001}$ & $0.566_{0.003}$ & $0.888_{0.000}$ & $0.882_{0.000}$ & $0.860_{0.001}$ & $0.544_{0.003}$ & $0.876_{0.001}$ & $0.882_{0.000}$ & $0.848_{0.001}$ & $0.527_{0.001}$ & $0.864_{0.001}$ \\
        \textbf{MTD} & $0.898_{0.002}$ & $0.909_{0.001}$ & $0.680_{0.004}$ & $0.922_{0.001}$ & $0.891_{0.001}$ & $0.889_{0.001}$ & $0.632_{0.003}$ & $0.903_{0.002}$ & $0.879_{0.003}$ & $0.863_{0.002}$ & $0.572_{0.007}$ & $0.880_{0.000}$ \\
        \textbf{SIP} & $0.900_{0.001}$ & $0.911_{0.000}$ & $0.687_{0.002}$ & $0.923_{0.000}$ & $0.891_{0.001}$ & $0.890_{0.001}$ & $0.634_{0.005}$ & $0.904_{0.001}$ & $0.882_{0.002}$ & $0.863_{0.001}$ & $0.570_{0.004}$ & $0.880_{0.001}$ \\
        \rowcolor{gray!20}
        \textbf{ADRL}  & $\bm{0.901_{0.002}}$ & $\bm{0.919_{0.001}}$ & $\bm{0.702_{0.004}}$ & $\bm{0.930_{0.001}}$ & $\bm{0.894_{0.001}}$ & $\bm{0.908_{0.001}}$ & $\bm{0.677_{0.005}}$ & $\bm{0.920_{0.001}}$ & $\bm{0.883_{0.002}}$ & $\bm{0.892_{0.002}}$ & $\bm{0.647_{0.006}}$ & $\bm{0.904_{0.002}}$ \\
        \hline 
    \end{tabular}
    \end{adjustbox}
    \caption{Comparison of different methods based on real NBA data.} \label{TableIV}
\end{table*}

  \begin{figure}[htbp]
    \centering
    \includegraphics[width=1\linewidth]{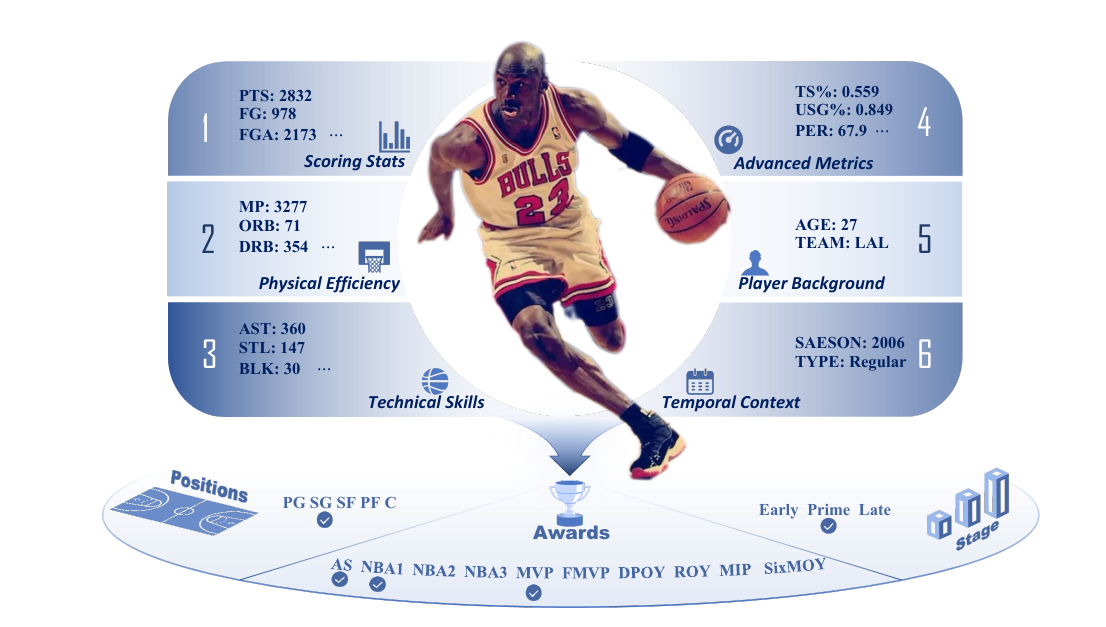}
    \caption{Description of the NBA dataset.}
    \label{fig:nba}
\end{figure}
\subsubsection{Construction of the Application Dataset}
The NBA dataset was sourced from publicly available records on Basketball-Reference \cite{nba-reference}, which spans all regular and playoff seasons from 2002 to 2022 and consists of 16,992 player-season instances. As shown in Fig. \ref{fig:nba},
each sample is described by features grouped into six  distinct views. The first four views capture various aspects of player performance, including 20 scoring-related variables, 14 indicators of rebounding and physical attributes, 15  metrics of technical statistics, and 10 measures of advanced efficiency. The remaining two views encode contextual information, with 41 variables representing player-level factors such as age and team affiliation, and 22 variables reflecting season-level attributes, including the year and playoff status. In addition, each instance is annotated with 18 labels that indicate career stage, playing position, and individual honors. Career progression is partitioned into early (first 25\%), peak (middle 50\%), and late (final 25\%) phases based on  each player's professional timeline.  Positions are represented using five one-hot variables corresponding to the standard roles. Additional multiple binary indicators denote major honors, such as MVP and Defensive Player of the Year, which provides a comprehensive assessment of each player's ability.

\subsubsection{Prediction Result Under Data Missing} To evaluate the predictive capability of our method under conditions of partial player features and sparse annotations, we conduct comparative experiments with missing ratios ranging from 50\% to 90\%. The  corresponding outcomes, as presented in Table \ref{TableIV}, demonstrate that our method consistently outperforms the baseline approaches across all evaluation metrics, regardless of the degree of missing data.  Moreover, our method achieves an AP of 0.647 in the case of 90\% data absence, whereas no other method surpasses 0.580, which highlights the notable advantage of our approach in this application. Even with incomplete statistical records, limited player information, and missing individual attributes,  our ADRL is still able to generate accurate predictions of player potential. This ability reveals  the robustness of our approach in adverse data environments and its strong potential for real-world applications.

\section{Conclusion}
In this paper, we propose a adaptive disentangled representation learning method (ADRL) for tackling  the iMvMLC problem. Specifically,  we perform view recovery by leveraging cross-view propagation of sample association information along with a neighborhood-aware weighting strategy, and further incorporate a random masking mechanism to suppress the influence of low-quality reconstructions. Moreover,  we develop a mutual information-constrained model to alleviate information bias during feature extraction and facilitate inter-view interaction within the shared feature space. Meanwhile, the model suppresses redundancy between view-specific features and external information, thereby enabling adaptive separation of shared and unique components. Label relevant information is propagated through a graph attention network to enhance the label distribution and induce semantically correlated label prototypes. Furthermore, matching these prototypes with  view representations produces a pseudo label space, whose structural information is used to guide efficient multi-view fusion and generate final prediction. Finally, the effectiveness of ADRL is  demonstrated through extensive experiments and its deployment on the NBA dataset.  Looking ahead, we plan to integrate the prior knowledge from large language models to facilitate  representation extraction and label semantic understanding.

\section{Acknowledgment}
This work was partially supported by the National Natural Science Foundation of China [Grant No. 62302516,62376281], and the Key NSF of China under Grant No. 62136005.

\bibliographystyle{IEEEtran}
\bibliography{bibtex}

@inproceedings{liu2023dicnet,
  title={Dicnet: Deep instance-level contrastive network for double incomplete multi-view multi-label classification},
  author={Liu, Chengliang and Wen, Jie and Luo, Xiaoling and Huang, Chao and Wu, Zhihao and Xu, Yong},
  booktitle={Proceedings of the AAAI conference on artificial intelligence},
  volume={37},
  number={7},
  pages={8807--8815},
  year={2023}
}

@inproceedings{tan2018incomplete,
  title={Incomplete multi-view weak-label learning.},
  author={Tan, Qiaoyu and Yu, Guoxian and Domeniconi, Carlotta and Wang, Jun and Zhang, Zili},
  booktitle={Ijcai},
  pages={2703--2709},
  year={2018}
}

@article{li2021concise,
  title={A concise yet effective model for non-aligned incomplete multi-view and missing multi-label learning},
  author={Li, Xiang and Chen, Songcan},
  journal={IEEE Transactions on Pattern Analysis and Machine Intelligence},
  volume={44},
  number={10},
  pages={5918--5932},
  year={2021},
  publisher={IEEE}
}

@inproceedings{duygulu2002object,
  title={Object recognition as machine translation: Learning a lexicon for a fixed image vocabulary},
  author={Duygulu, Pinar and Barnard, Kobus and de Freitas, Joao FG and Forsyth, David A},
  booktitle={Computer vision—ECCV 2002: 7th European conference on computer vision copenhagen, Denmark, May 28--31, 2002 proceedings, part IV 7},
  pages={97--112},
  year={2002},
  organization={Springer}
}

@article{everingham2010pascal,
  title={The pascal visual object classes (voc) challenge},
  author={Everingham, Mark and Van Gool, Luc and Williams, Christopher KI and Winn, John and Zisserman, Andrew},
  journal={International journal of computer vision},
  volume={88},
  pages={303--338},
  year={2010},
  publisher={Springer}
}

@article{sun2024deep,
  title={A Deep Model for Partial Multi-label Image Classification with Curriculum-based Disambiguation},
  author={Sun, Feng and Xie, Ming-Kun },
  journal={Machine Intelligence Research},
  pages={1--14},
  year={2024},
  publisher={Springer}
}

@inproceedings{chang2020taming,
  title={Taming pretrained transformers for extreme multi-label text classification},
  author={Chang, Wei-Cheng and Yu, Hsiang-Fu and Zhong, Kai and Yang, Yiming and Dhillon, Inderjit S},
  booktitle={Proceedings of the 26th ACM SIGKDD International Conference on knowledge Discovery and Data Mining},
  pages={3163--3171},
  year={2020}
}

@inproceedings{li2024deep,
  title={Deep incomplete multi-view network semi-supervised multi-label learning with unbiased loss},
  author={Li, Quanjiang and Luo, Tingjin and Jiang, Mingdie and Liao, Jiahui and Jiang, Zhangqi},
  booktitle={Proceedings of the 32nd ACM International Conference on Multimedia},
  pages={9048--9056},
  year={2024}
}

@inproceedings{hao2025uncertainty,
  title={Uncertainty-Aware Global-View Reconstruction for Multi-View Multi-Label Feature Selection},
  author={Hao, Pingting and Liu, Kunpeng and Gao, Wanfu},
  booktitle={Proceedings of the AAAI Conference on Artificial Intelligence},
  volume={39},
  number={16},
  pages={17068--17076},
  year={2025}
}

@ARTICLE{liu2025reliable,
  author={Liu, Chengliang and Wen, Jie and Xu, Yong and Zhang, Bob and Nie, Liqiang and Zhang, Min},
  journal={IEEE Transactions on Pattern Analysis and Machine Intelligence}, 
  title={Reliable Representation Learning for Incomplete Multi-View Missing Multi-Label Classification}, 
  year={2025},
  volume={},
  number={},
  pages={1-17},
}

@inproceedings{liu2015low,
  title={Low-rank multi-view learning in matrix completion for multi-label image classification},
  author={Liu, Meng and Luo, Yong and Tao, Dacheng and Xu, Chao and Wen, Yonggang},
  booktitle={Proceedings of the AAAI conference on artificial intelligence},
  volume={29},
  number={1},
  year={2015}
}

@article{hao2025embedded,
  title={Embedded feature fusion for multi-view multi-label feature selection},
  author={Hao, Pingting and Gao, Wanfu and Hu, Liang},
  journal={Pattern Recognition},
  volume={157},
  pages={110888},
  year={2025},
  publisher={Elsevier}
}

@ARTICLE{SEC-LSRM,
  author={Guan, Renxiang and Liu, Tianrui and Tu, Wenxuan and Tang, Chang and Luo, Wenhan and Liu, Xinwang},
  journal={IEEE Transactions on Knowledge and Data Engineering}, 
  title={Sampling Enhanced Contrastive Multi-View Remote Sensing Data Clustering with Long-Short Range Information Mining}, 
  year={2025},
  volume={},
  number={},
  pages={1-15}
}

@inproceedings{sun2024fedmlp,
  title={Fedmlp: Federated multi-label medical image classification under task heterogeneity},
  author={Sun, Zhaobin and Wu, Nannan and Shi, Junjie and Yu, Li and Cheng, Kwang-Ting and Yan, Zengqiang},
  booktitle={International Conference on Medical Image Computing and Computer-Assisted Intervention},
  pages={394--404},
  year={2024},
  organization={Springer}
}

@inproceedings{liu2023incomplete,
  title={Incomplete multi-view multi-label learning via label-guided masked view-and category-aware transformers},
  author={Liu, Chengliang and Wen, Jie and Luo, Xiaoling and Xu, Yong},
  booktitle={Proceedings of the AAAI conference on artificial intelligence},
  volume={37},
  number={7},
  pages={8816--8824},
  year={2023}
}

@inproceedings{liu2024partial,
  title={Partial multi-view multi-label classification via semantic invariance learning and prototype modeling},
  author={Liu, Chengliang and Xu, Gehui and Wen, Jie and Liu, Yabo and Huang, Chao and Xu, Yong},
  booktitle={Forty-first International Conference on Machine Learning},
  year={2024}
}

@inproceedings{liu2024attention,
  title={Attention-induced embedding imputation for incomplete multi-view partial multi-label classification},
  author={Liu, Chengliang and Jia, Jinlong and Wen, Jie and Liu, Yabo and Luo, Xiaoling and Huang, Chao and Xu, Yong},
  booktitle={Proceedings of the AAAI Conference on Artificial Intelligence},
  volume={38},
  number={12},
  pages={13864--13872},
  year={2024}
}

@article{liu2023masked,
  title={Masked two-channel decoupling framework for incomplete multi-view weak multi-label learning},
  author={Liu, Chengliang and Wen, Jie and Liu, Yabo and Huang, Chao and Wu, Zhihao and Luo, Xiaoling and Xu, Yong},
  journal={Advances in Neural Information Processing Systems},
  volume={36},
  pages={32387--32400},
  year={2023}
}

@article{xiao2024new,
  title={A new multi-view multi-label model with privileged information learning},
  author={Xiao, Yanshan and Chen, Junfeng and Liu, Bo and Zhao, Liang and Kong, Xiangjun and Hao, Zhifeng},
  journal={Information Sciences},
  volume={656},
  pages={119911},
  year={2024},
  publisher={Elsevier}
}

@inproceedings{von2004labeling,
  title={Labeling images with a computer game},
  author={Von Ahn, Luis and Dabbish, Laura},
  booktitle={Proceedings of the SIGCHI conference on Human factors in computing systems},
  pages={319--326},
  year={2004}
}

@inproceedings{grubinger2006iapr,
  title={The iapr tc-12 benchmark: A new evaluation resource for visual information systems},
  author={Grubinger, Michael and Clough, Paul and M{\"u}ller, Henning and Deselaers, Thomas},
  booktitle={International workshop ontoImage},
  volume={2},
  year={2006}
}

@inproceedings{huiskes2008mir,
  title={The mir flickr retrieval evaluation},
  author={Huiskes, Mark J and Lew, Michael S},
  booktitle={Proceedings of the 1st ACM international conference on Multimedia information retrieval},
  pages={39--43},
  year={2008}
}

@article{hao2024anchor,
  title={Anchor-guided global view reconstruction for multi-view multi-label feature selection},
  author={Hao, Pingting and Liu, Kunpeng and Gao, Wanfu},
  journal={Information Sciences},
  volume={679},
  pages={121124},
  year={2024},
  publisher={Elsevier}
}

@article{ma2021expand,
  title={Expand globally, shrink locally: Discriminant multi-label learning with missing labels},
  author={Ma, Zhongchen and Chen, Songcan},
  journal={Pattern Recognition},
  volume={111},
  pages={107675},
  year={2021},
  publisher={Elsevier}
}

@article{zhao2022non,
  title={Non-aligned multi-view multi-label classification via learning view-specific labels},
  author={Zhao, Dawei and Gao, Qingwei and Lu, Yixiang and Sun, Dong},
  journal={IEEE Transactions on Multimedia},
  volume={25},
  pages={7235--7247},
  year={2022},
  publisher={IEEE}
}

@article{wen2023deep,
  title={Deep double incomplete multi-view multi-label learning with incomplete labels and missing views},
  author={Wen, Jie and Liu, Chengliang and Deng, Shijie and Liu, Yicheng and Fei, Lunke and Yan, Ke and Xu, Yong},
  journal={IEEE transactions on neural networks and learning systems},
  year={2023},
  publisher={IEEE}
}

@inproceedings{zhang2018latent,
  title={Latent semantic aware multi-view multi-label classification},
  author={Zhang, Changqing and Yu, Ziwei and Hu, Qinghua and Zhu, Pengfei and Liu, Xinwang and Wang, Xiaobo},
  booktitle={Proceedings of the AAAI conference on artificial intelligence},
  volume={32},
  number={1},
  year={2018}
}

@article{zhao2021consistency,
  title={Consistency and diversity neural network multi-view multi-label learning},
  author={Zhao, Dawei and Gao, Qingwei and Lu, Yixiang and Sun, Dong and Cheng, Yusheng},
  journal={Knowledge-Based Systems},
  volume={218},
  pages={106841},
  year={2021},
  publisher={Elsevier}
}

@inproceedings{wu2019multi,
  title={Multi-View Multi-Label Learning with View-Specific Information Extraction.},
  author={Wu, Xuan and Chen, Qing-Guo and Hu, Yao and Wang, Dengbao and Chang, Xiaodong and Wang, Xiaobo and Zhang, Min-Ling},
  booktitle={IJCAI},
  pages={3884--3890},
  year={2019}
}

@article{liu2022localized,
  title={Localized sparse incomplete multi-view clustering},
  author={Liu, Chengliang and Wu, Zhihao and Wen, Jie and Xu, Yong and Huang, Chao},
  journal={IEEE Transactions on Multimedia},
  volume={25},
  pages={5539--5551},
  year={2022},
  publisher={IEEE}
}

@inproceedings{wen2019unified,
  title={Unified embedding alignment with missing views inferring for incomplete multi-view clustering},
  author={Wen, Jie and Zhang, Zheng and Xu, Yong and Zhang, Bob and Fei, Lunke and Liu, Hong},
  booktitle={Proceedings of the AAAI conference on artificial intelligence},
  volume={33},
  number={01},
  pages={5393--5400},
  year={2019}
}

@article{zhu2017multi,
  title={Multi-label learning with global and local label correlation},
  author={Zhu, Yue and Kwok, James T and Zhou, Zhi-Hua},
  journal={IEEE Transactions on Knowledge and Data Engineering},
  volume={30},
  number={6},
  pages={1081--1094},
  year={2017},
  publisher={IEEE}
}

@inproceedings{chen2019learning,
  title={Learning semantic-specific graph representation for multi-label image recognition},
  author={Chen, Tianshui and Xu, Muxin and Hui, Xiaolu and Wu, Hefeng and Lin, Liang},
  booktitle={Proceedings of the IEEE/CVF international conference on computer vision},
  pages={522--531},
  year={2019}
}

@article{xu2022self,
  title={Self-supervised discriminative feature learning for deep multi-view clustering},
  author={Xu, Jie and Ren, Yazhou and Tang, Huayi and Yang, Zhimeng and Pan, Lili and Yang, Yang and Pu, Xiaorong and Yu, Philip S and He, Lifang},
  journal={IEEE Transactions on Knowledge and Data Engineering},
  volume={35},
  number={7},
  pages={7470--7482},
  year={2022},
  publisher={IEEE}
}

@article{wen2022survey,
  title={A survey on incomplete multiview clustering},
  author={Wen, Jie and Zhang, Zheng and Fei, Lunke and Zhang, Bob and Xu, Yong and Zhang, Zhao and Li, Jinxing},
  journal={IEEE Transactions on Systems, Man, and Cybernetics: Systems},
  volume={53},
  number={2},
  pages={1136--1149},
  year={2022},
  publisher={IEEE}
}

@article{fang2021animc,
  title={Animc: A soft approach for autoweighted noisy and incomplete multiview clustering},
  author={Fang, Xiang and Hu, Yuchong and Zhou, Pan and Wu, Dapeng},
  journal={IEEE Transactions on Artificial Intelligence},
  volume={3},
  number={2},
  pages={192--206},
  year={2021},
  publisher={IEEE}
}

@article{fang2022multi,
  title={Multi-modal cross-domain alignment network for video moment retrieval},
  author={Fang, Xiang and Liu, Daizong and Zhou, Pan and Hu, Yuchong},
  journal={IEEE Transactions on Multimedia},
  volume={25},
  pages={7517--7532},
  year={2022},
  publisher={IEEE}
}

@inproceedings{jiang2024deep,
  title={Deep Incomplete Multi-View Learning Network with Insufficient Label Information},
  author={Jiang, Zhangqi and Luo, Tingjin and Liang, Xinyan},
  booktitle={Proceedings of the AAAI Conference on Artificial Intelligence},
  volume={38},
  number={11},
  pages={12919--12927},
  year={2024}
}

@INPROCEEDINGS{9879206,
  author={He, Kaiming and Chen, Xinlei and Xie, Saining and Li, Yanghao and Dollár, Piotr and Girshick, Ross},
  booktitle={IEEE/CVF Conference on Computer Vision and Pattern Recognition}, 
  title={Masked Autoencoders Are Scalable Vision Learners}, 
  year={2022},
  volume={},
  number={},
  pages={15979-15988},
  doi={10.1109/CVPR52688.2022.01553}}

@inproceedings{LiLL25,
  author       = {Quanjiang Li and
                  Tingjin Luo and
                  Jiahui Liao},
  title        = {Theory-Inspired Deep Multi-View Multi-Label Learning with Incomplete
                  Views and Noisy Labels},
  booktitle    = {IEEE/CVF Conference on Computer Vision and Pattern Recognition},
  pages        = {20706--20715},
  publisher    = {Computer Vision Foundation / {IEEE}},
  year         = {2025},
}

@inproceedings{45903,
title	= {Deep Variational Information Bottleneck},
author	= {Alex Alemi and Ian Fischer and Josh Dillon and Kevin Murphy},
year	= {2017},
URL	= {https://arxiv.org/abs/1612.00410},
booktitle	= {
International Conference on Learning Representations}}

@inproceedings{bao2021disentangled,
  title={Disentangled variational information bottleneck for multiview representation learning},
  author={Bao, Feng},
  booktitle={CAAI International Conference on Artificial Intelligence},
  pages={91--102},
  year={2021},
  organization={Springer}
}

@article{lian2023online,
  title={Online learning from evolving feature spaces with deep variational models},
  author={Lian, Heng and Wu, Di and Hou, Bo-Jian and Wu, Jian and He, Yi},
  journal={IEEE Transactions on Knowledge and Data Engineering},
  volume={36},
  number={8},
  pages={4144--4162},
  year={2023},
  publisher={IEEE}
}

@article{huang2023generalized,
  title={Generalized information-theoretic multi-view clustering},
  author={Huang, Weitian and Yang, Sirui and Cai, Hongmin},
  journal={Advances in Neural Information Processing Systems},
  volume={36},
  pages={58752--58764},
  year={2023}
}

@inproceedings{li2025semi,
  title={Semi-Supervised Multi-View Multi-Label Learning with View-Specific Transformer and Enhanced Pseudo-Label},
  author={Li, Quanjiang and Luo, Tingjin and Jiang, Mingdie and Jiang, Zhangqi and Hou, Chenping and Li, Feijiang},
  booktitle={Proceedings of the AAAI Conference on Artificial Intelligence},
  volume={39},
  number={17},
  pages={18430--18438},
  year={2025}
}

@misc{nba-reference,  
  title        = {NBA Player Totals Statistics},   
  author       = { {Basketball Reference} },  
  year         = {2025},
  note         = {Accessed: 2025-03-03}, 
  howpublished = {\url{https://www.basketball-reference.com}},
}
\vspace{2cm}
\begin{IEEEbiographynophoto}{Quanjiang Li}
is a M.S. candidate at the National University of Defense Technology, Changsha, China. He received the B.S. degree from National University of Defense Technology in 2023. His research interests include data mining and machine learning.
\end{IEEEbiographynophoto}

\vspace{-9cm}

\begin{IEEEbiographynophoto}{Zhiming Liu} is an undergraduate student at Harbin Institute of Technology, Shenzhen, China, where he enrolled in 2023. His research interests include machine learning and transfer learning.
\end{IEEEbiographynophoto}

\vspace{-9cm}

\begin{IEEEbiographynophoto}{Tianxiang Xu} received his B.S. degree in Cyberspace Security from the University of Jinan in 2024. He is currently pursuing an M.S. degree at the School of Software and Microelectronics, Peking University. His research interests lie in artificial intelligence, with a particular focus on large language models and cybersecurity.
\end{IEEEbiographynophoto}
\vspace{-9cm}
\begin{IEEEbiographynophoto}{Tingjin Luo} received the B.S., Master and Ph.D degrees from the National University of Defense Technology, Changsha, China. He is currently an Associate Professor with the College of Science of the same university. He has authored more than 50 papers in journals and conferences, such as IEEE TPAMI, IEEE TKDE, IEEE TCYB, IEEE TIP, CVPR and ACM KDD etc. He has been a Program Committee member of several conferences including ICML, CVPR, IJCAI, AAAI and ICPR etc. His research interests include weakly supervised learning, multi-view learning and data mining. 
\end{IEEEbiographynophoto}

\vspace{-9cm}

\begin{IEEEbiographynophoto}{Chenping Hou}
(Member, IEEE) received the BS and PhD degrees in applied mathematics from the National University of Defense Technology, Changsha, China, in 2004 and 2009, respectively. He is currently a full professor with the College of Science, National University of Defense Technology. He has authored more than 100 papers in journals and conferences, such as the IEEE Transactions on Pattern Analysis and Machine Intelligence, IEEE Transactions on Knowledge and Data Engineering, IEEE Transactions on Neural Networks and Learning Systems, IEEE Transactions on Cybernetics, IEEE Transactions on Image Processing, Pattern Recognition, IJCAI, AAAI, etc. He has been a Program Committee member of several conferences including IJCAI, AAAI, etc. His current research interests include pattern recognition, machine learning, data mining, and computer vision.
\end{IEEEbiographynophoto}

\end{document}